\documentclass[10pt,twocolumn,letterpaper]{article}

\usepackage[pagenumbers]{cvpr} 

\usepackage[dvipsnames]{xcolor}

\usepackage{adjustbox}
\usepackage{graphicx}
\usepackage{amsmath}
\usepackage{amssymb}
\usepackage{booktabs}
\usepackage{soul,color,xcolor}
\usepackage{algorithm}
\usepackage{algpseudocode}
\usepackage{array}
\usepackage{multirow}

\definecolor{cvprblue}{rgb}{0.21,0.49,0.74}
\usepackage[pagebackref,breaklinks,colorlinks,citecolor=cvprblue]{hyperref}

\def\paperID{11592} 
\def\confName{CVPR}
\def\confYear{2024}

\title{Linear Relative Pose Estimation Founded on Pose-only Imaging Geometry}

\author{Qi Cai\textsuperscript{1}, Xinrui Li\textsuperscript{1}, Yuanxin Wu\textsuperscript{1,\thanks{Corresponding author}}\\ 
\textsuperscript{1}Shanghai Jiao Tong University\\
{\tt\small qicaiCN@gmail.com, physalis@sjtu.edu.cn, yuanx\_wu@hotmail.com}
}

\begin{document}

\maketitle
\begin{abstract}
How to efficiently and accurately handle image matching outliers is a critical issue in two-view relative estimation. The prevailing RANSAC method necessitates that the minimal point pairs be inliers. This paper introduces a linear relative pose estimation algorithm for n $( n \geq 6$) point pairs, which is founded on the recent pose-only imaging geometry to filter out outliers by proper reweighting.  The proposed algorithm is able to handle planar degenerate scenes, and enhance robustness and accuracy in the presence of a substantial ratio of outliers. Specifically, we embed the linear global translation (LiGT) constraint into the strategies of iteratively reweighted least-squares (IRLS) and RANSAC so as to realize robust outlier removal. Simulations and real tests of the Strecha dataset show that the proposed algorithm achieves relative rotation accuracy improvement of 2 $\sim$ 10 times in face of as large as 80\% outliers.
\end{abstract}    
\section{Introduction}
\label{sec:intro}
The popular pipeline in three-dimension (3D) visual computing, such as simultaneous localization and mapping (SLAM) and structure-from-motion (SfM), involves~\cite{ozyecsil2017survey,szeliski2022computer}: 1) extracting, matching, and tracking image features; 2) removing image matching outliers through visual geometric constraints, such as the fundamental/essential/homography matrix equation; 3) estimating camera poses; and 4) reconstructing the 3D scene. While image feature extraction methods employ image Laplacian operators (like the SIFT~\cite{lowe2004distinctive}, ORB~\cite{rublee2011orb} or SURF~\cite{bay2008speeded}) to identify interest points, matching outliers often occur in challenging scenarios such as those with insufficient lighting, occlusions, moving objects, and repetitive textures~\cite{Jing2023TPAMI_survey}. Those outliers would compromise the essential matrix~\cite{longuet1981computer} solution to two-view relative pose estimation. Additionally, when solving the essential equation, the two-view pose estimation encounters degeneracy problems~\cite{decker2008dealing,nister2004efficient,stewenius2006recent,kneip2012finding}  (i.e. pure rotational motion or planar structure).  
 
Unfortunately, ensuring robustness to outliers remains a big challenge to 3D visual computing. For example, although the five-point algorithm can deal with the planar scene and pure rotational motion, it utilizes exactly five point pairs to construct five independent constraints of the essential matrix equation and to identify the true solution in the four-dimensional solution space~\cite{nister2004efficient,stewenius2006recent} by using an action matrix~\cite{stewenius2006recent}. As a matter of fact,  the essential equation under the planar scene or pure rotational motion has six independent constraints on the essential matrix for $ n \geq 6 $ point pairs~\cite{Philip1998,pizarro2003relative}. Consequently, the five-point algorithm lacks the constraint information of the essential matrix and is likely subject to multiple solutions among which the right one needs to identify (\eg by using the Sampson distance of the co-planar constraint~\cite{hartley2003multiple}).

Robust estimation methods such as RANSAC~\cite{fischler1981random}, LMedS~\cite{rousseeuw1984least}, and M-estimation~\cite{huber1992robust,huber2004robust} are essential to screen ubiquitous outliers. Specifically, RANSAC utilizes random sampling and consistency checks, combined with geometric constraints such as coplanar equations, to segregate inliers from outliers and to estimate the essential, fundamental or homography matrix. In the CVPR 2023 highlight paper, Peng \etal~\cite{peng2023convergence} introduced the GNC-IRLS method, a graduated non-convex strategy (GNC) for iteratively reweighted least-squares (IRLS), showcasing rapid convergence and accuracy in the context of 3D point registration. 

The recent work~\cite{cai2021pose} proposed a promising pose-only imaging geometry, which is equivalent to the classical multiple-view geometry and can be equivalently described using the linear global translation (LiGT) constraint. This pose-only representation has a significantly lower parameter space dimensionality, in contrast to the conventional bundle adjustment, by decoupling the 3D scene from the camera pose. Notably, the LiGT algorithm was embedded into the core library of OpenMVG~\cite{moulon2017openmvg} for global translation estimation, which is able to linearly solve global translations on given rotations of all views. It can handle some specific degenerate motions, such as local co-linearity, parallel rigidity and pure rotational motion~\cite{cai2021pose}. 

The primary contributions of this paper include:

1. Propose a linear relative pose estimation algorithm (LiRP) based on the essential matrix equation to process all point pairs in one batch, enhancing the estimation accuracy and the capability of handling the degenerate planar scene and pure rotational motion.
   
2. Introduce the pose-only imaging geometry into the LiRP to screen ouliers.  The outlier identification is greatly improved by the re-weighted LiGT constraint's residual thanks to the compete property of pose-only imaging geometry.

3. Incorporate RANSAC and the recent GNC-IRLS method into the LiRP for further robustness improvement.

The paper is organized as follows. \cref{sec:related works} discusses the related works on two-view robust pose estimation. \cref{sec:Preliminaries and Notation} presents the predefined mathematical notations. \cref{sec:Two-view Constraints and Residuals} reviews the constraints in two-view geometry and their residual statistics. The LiRP algorithm is proposed in \cref{sec:LiRP}. \cref{sec:GNC-RANSAC method} elaborates on the combination of GNC-IRLS with RANSAC. \cref{sec:Experiments} presents both simulation and real data test results. The paper is concluded in \cref{sec:conclusion}.
\section{Related Works}
\label{sec:related works}
\subsection{Robust Estimation}

In 1981, Fischler and Bolles introduced the RANSAC method~\cite{fischler1981random}, which subsequently became the mainstream outlier handling method in SLAM and SfM. The RANSAC method requires empirical parameters to be set according to specific estimation problems, such as the proportion of inliers to outliers and outlier detection error thresholds. Rousseeuw~\cite{rousseeuw1984least} introduced the LMedS method, which selects the optimal sub-sample using the minimum median deviation criterion with no need of presetting the threshold to distinguish inliers from outliers. The AC-RANSAC framework by Moisan \etal~\cite{moulon2013adaptive} adaptively updates RANSAC's inlier and outlier parameters. Nonetheless, in 2017, Özyeşil \etal pointed out in~\cite{ozyecsil2017survey} that the use of RANSAC could reduce the outlier mismatch rate to a higher yet acceptable level, for instance, 40\% in offline applications; however, in real-time applications, the outlier ratio could be as high as 80\% $\sim$ 90\%.

Furthermore, the robust M-estimator method~\cite{huber1992robust,de2021review} is often employed in 3D vision such as pose graphs and bundle adjustments. It utilizes a loss function to weight the cost function, and aims to obtain robust solutions when faced with anomalies or non-normal data distributions. In two-view pose estimation, for example, a recent study~\cite{zhao2020efficient} proposed a convex optimization-based solver for essential matrix estimation using all point pairs, which utilizes the M-estimator to handle outliers. 

Another prevalent method for outlier treatment is based on the IRLS~\cite{holland1977robust,huber2004robust}, which reweights the conventional least squares, highly susceptible to outliers, by using the observation residual. For instance, in global rotaion averaging, Chatterjee \etal~\cite{chatterjee2017robust} utilized the IRLS scheme to enhance the estimation robustness. The recent study~\cite{peng2023convergence} proved the convergence of GNC-IRLS for outlier-robust estimation and proposed a smooth majorizer function and superlinear schedule update rule for IRLS.

\subsection{Two-view Pose Constraints and Estimation}

Nowadays, the mainstream methodologies of two-view relative pose estimation , for example those in such popular platforms as OpenMVG~\cite{moulon2017openmvg}, COLMAP~\cite{schonberger2016structure} and OpenGV~\cite{kneip2014opengv}, unexceptionally rely on the minimal matching point pairs and the essential matrix equation~\cite{hartley2003multiple,longuet1981computer,stewenius2006recent,kneip2014opengv,moulon2017openmvg}. The advantage of the essential matrix coplanar equation is that it allows for a linear solution to the relative pose, but it loses the depth information and is an incomplete representation of two-view geometry~\cite{cai2019equivalent,cai2021pose}. In fact, the relative pose estimation based on the essential matrix equation faces a number of challenges, such as multiple solutions, planar scenes, and pure rotational motion, among others~\cite{nister2004efficient,stewenius2006recent,pizarro2003relative,Philip1998,kneip2012finding,zhao2020efficient}. In 1996, Philip introduced a non-iterative algorithm to determine all essential matrices corresponding to five point pairs~\cite{philip1996non} and proposed a linear method for solving the essential matrix from six point pairs. In 2003, Pizarro \etal~\cite{pizarro2003relative} introduced a six-point algorithm for robustly estimating the essential matrix, which remains effective even in the context of planar scenes. In 2004, Nister~\cite{nister2004efficient} employed the approach of solving the underdetermined group of coplanar equations to analyze the relationship between ternary polynomials and proposed a five-point algorithm for solving the essential matrix. In 2006, Stevenius and Nister~\cite{stewenius2006recent} incorporated the Grobner base theory to further advance the five-point method.

Kneip \etal~\cite{kneip2012finding} believed that the traditional coplanar constraint had issues in expressing pure rotation, and then constructed a general coplanar constraint to describe two-view geometry. In specific, the authors utilized the Grobner base to solve the relative pose using the five-point method, and provided a method to identify the correct pose solution without 3D reconstruction. However, the shortcomings of the coplanar equation (loss of depth information and the presence of multiple solutions) has not been addressed until the work by Cai \etal~\cite{cai2019equivalent} in 2019. The work introduced a pair of two-view pose-only constraints equivalent to the two-view imaging relationship, analytically decoupling the camera pose from the 3D point. 
Afterwards, the authors further extended the pose-only representation to multi-view imaging case, discovered a linear relationship between depth expression and translation, and came up with the LiGT constraint equivalent to the classical multi-view geometry~\cite{cai2021pose}.

\section{ Preliminaries and Notation}
\label{sec:Preliminaries and Notation}
In this paper, we use the tilde symbol to denote the bearing vector form of a vector,  i.e., $\boldsymbol{\tilde{a}}=\boldsymbol{a}/\| \boldsymbol{a} \|$, where $\| 
\boldsymbol{a} \|$ denotes the norm of the vector $\boldsymbol{a}$.
The relative pose between the left and right views is  represented by $R$ and $\boldsymbol{t}$, respectively. The skew-symmetric matrix formed by any vector $\boldsymbol{a}$ is denoted as $[\boldsymbol{a}]_{\times}$. 

Assume that the left and right views correspond to camera coordinate systems $C$ and $C^{\prime}$ centered at $c$ and $c'$, respectively. Consider $n$ 3D feature points, where $\boldsymbol{X}_i^C=\left(x_i^C, y_i^C, z_i^C\right)^T$ in the left camera system and $\boldsymbol{X}_i^{ C^{\prime}}=\left(x_i^{C^{\prime}}, y_i^{C^{\prime}}, z_i^{C^{\prime}}\right)^T$ in the right, the normalized image coordinates of the 3D point projected onto the left and right views are $\boldsymbol{x}_i=\boldsymbol{X}_i^C / z_i^C=\left(x_i, y_i, 1\right)^T$ and $\boldsymbol{x}_i^{\prime}=\boldsymbol{X}_i^{C^{\prime}} / z_i^{C^{\prime}}=\left(x_i^{\prime}, y_i^{\prime}, 1\right)^T$,  for $i=1,\ldots ,n$. Here, $z_i^C$ and $z_i^{C^{\prime}}$ respectively represent the depths of the 3D points in the left and right views~\cite{cai2021pose}. ${\boldsymbol{x}_{i}}$ and ${\boldsymbol x_i^{\prime}}$ denote the normalized image coordinates of the corresponding point pair. 

\section{Two-view Constraints and Residuals}
\label{sec:Two-view Constraints and Residuals}
In the community of computer vision, the two-view geometry concerns the interrelation between two images of a specified scene, taken from two viewpoints. The two-view geometry is dictated by multiple different constraints that are crucial to a plethora of computer vision endeavors.

The 3D feature point in the two views are related via their relative pose by:
\begin{equation}
\boldsymbol{X}_i^{C^{\prime}} = R \boldsymbol{X}_i^C + \boldsymbol{t}.
\label{eq:eq1}
\end{equation}
Then, the classical two-view imaging equation can be expressed as:
\begin{equation}
z_i^{C^{\prime}} \boldsymbol{x}_i^{\prime} = z_i^C R \boldsymbol{x}_i + \boldsymbol{t} \Leftrightarrow \boldsymbol{x}_i^{\prime} = \lambda_i (R \boldsymbol{x}_i + s_i \boldsymbol{t}),
\label{eq:eq2}
\end{equation}
where $\lambda_i \triangleq z_i^C / z_i^{C^{\prime}} \in \mathbb{R}^{+}$ and $s_i \triangleq 1 / z_i^C \in \mathbb{R}^{+}$ represent unknown depth factors. Left-multiplying the equation on both sides by $\boldsymbol x_i^{\prime T}[\boldsymbol t]_{\times}$ to eliminate the depth factors, then we obtain the Longuet-Higgins's coplanar constraint~\cite{longuet1981computer}:
\begin{equation}
\boldsymbol{0} = \boldsymbol{x}_i^{\prime T}[\boldsymbol{t}]_{\times} \boldsymbol{x}_i^{\prime} = \lambda_i \boldsymbol{x}_i^{\prime T}[\boldsymbol{t}]_{\times} R \boldsymbol{x}_i
\Leftrightarrow \boldsymbol{x}_i^{\prime T} E \boldsymbol{x}_i = 0,
\label{eq:eq4}
\end{equation}
The derivation from \cref{eq:eq2} to \cref{eq:eq4} is irreversible, and it should noted that \cref{eq:eq4} loses partial geometric imaging information, such as the forward intersection of projection rays~\cite{cai2021pose}. In 2012, Kneip \etal~\cite{kneip2012finding} proposed a general coplanar equation to describe the relationship in two-view imaging geometry. It can be obtained by left-multiplying \cref{eq:eq2} by $\boldsymbol{t}^T\left[\boldsymbol{x}_i^{\prime}\right]_\times$, that is:
\begin{equation}
\begin{aligned}
& \boldsymbol{0} = \lambda_i \boldsymbol{t}^T\left[\boldsymbol{x}_i^{\prime}\right]_{\times}(R \boldsymbol{x}_i + s_i \boldsymbol{t}) = \lambda_i \boldsymbol{t}^T\left[\boldsymbol{x}_i^{\prime}\right]_{\times} R \boldsymbol{x}_i \\
& \Leftrightarrow \boldsymbol{t}^T(\boldsymbol{x}_i^{\prime} \times R \boldsymbol{x}_i) = \boldsymbol{0},
\end{aligned}
\label{eq:eq5}
\end{equation}
which means that all vectors $\{\boldsymbol{x}_i^{\prime} \times R \boldsymbol{x}_i | i = 1, ..., n\}$ are located on a plane with the relative translation $\boldsymbol t$ parallel to its normal vector. Define $B = (\boldsymbol{x}_1^{\prime} \times R \boldsymbol{x}_1, ..., \boldsymbol{x}_n^{\prime} \times R \boldsymbol{x}_n)^T$ and $M = B B^T$. According to \cref{eq:eq5}, the $B$ matrix is rank-deficient, i.e., the smallest eigenvalue of the $M$ matrix satisfies the constraint ~\cite{kneip2012finding,kneip2013direct}:
\begin{equation}
\lambda_M(R) = 0,
\label{eq:eq6}
\end{equation}
Due to the irreversibility of the above derivation, the constraint still loses some geometric information. Furthermore, it can be shown that the Kneip's equation is equivalent to the essential equation \cref{eq:eq4}, as detailed in the Appendix.A. Based on the coplanar equation constraint \cref{eq:eq4}, the residual for a matched point pair is constructed in the form of bearing vectors as
\begin{equation}
v^{E}_i(R, \boldsymbol t,\tilde{\boldsymbol x}_i, \tilde{\boldsymbol x}^{\prime}_i) = ||\tilde{\boldsymbol{x}}_i^{\prime T} E \tilde{\boldsymbol{x}}_i||.
\end{equation}
Unless otherwise specified, the residual vector formed by all point pairs will be denoted as $\boldsymbol v$. Kneip introduced a constraint for $R$ optimization with $\lambda_M(R)=0$ as shown in \cref{eq:eq6}. However, the constraint in the form of smallest eigenvalue cannot be readily used to construct a residual for a point pair. In view of the fact that it is proven to be equivalent to the coplanar equation, only $\boldsymbol v^{E}$ will be further explored in the sequel. Note that the two-view imaging equation in \cref{eq:eq2} contains unknown depths, and thus it is not feasible to directly construct a residual either. Nonetheless, if the relative pose is determined, the gold-standard bundle adjustment error in two-view geometry can be formed by re-projection, that is,
\begin{equation}
v_i^{BA}(R,\boldsymbol t, \boldsymbol X^W_i, {\boldsymbol x}_i,{\boldsymbol x}^{\prime}_i) =||\frac{\boldsymbol{\epsilon}_i} {\boldsymbol{e}_3^T \boldsymbol{\epsilon}_i}-{{\boldsymbol x}}_i^{\prime}||,
\end{equation}
where $\boldsymbol{\epsilon}_i = R(\boldsymbol X^W_i-\boldsymbol t)$, $\boldsymbol X^W_i$ is triangulated by the relative pose, and $\boldsymbol{e}_3^T$ is the third row of the identity matrix. Using the bearing vector, another reprojection residual used by OpenGV~\cite{kneip2014opengv} is
\begin{equation}
v_i^{OpenGV}(R, \boldsymbol t, \boldsymbol X^W_i, \tilde{\boldsymbol x}_i, \tilde{\boldsymbol x}^{\prime}_i) = ||1 - \tilde{\boldsymbol \epsilon}_i^T \tilde{\boldsymbol x}_i^{\prime}||.
\end{equation}
Recent studies~\cite{cai2019equivalent,cai2021pose} proposed the pose-only imaging geometry by representing the depth factors as functions of poses, which is provably equivalent to the classical multi-view geometry. According to~\cite{cai2019equivalent}, the pairwise pose-only (PPO) constraint is given as
\begin{equation}
\frac{\left\|\boldsymbol{t} \times R \boldsymbol{x}_i\right\|}{\left\|\boldsymbol{x}_i^{\prime} \times R \boldsymbol{x}_i\right\|} \boldsymbol{x}_i^{\prime}=\frac{\left\|\boldsymbol{t} \times \boldsymbol{x}_i^{\prime}\right\|}{\left\|\boldsymbol{x}_i^{\prime} \times R \boldsymbol{x}_i\right\|} R \boldsymbol{x}_i+\boldsymbol{t}.
\label{eq:eq7}
\end{equation}
It should be highlighted that the depth factors are linearly related to translation~\cite{cai2021pose}, namely, 
\begin{equation}
\begin{aligned}
& \frac{\left\|\boldsymbol{t} \times \boldsymbol{x}_i^{\prime}\right\|}{\left\|\boldsymbol{x}_i^{\prime} \times R \boldsymbol{x}_i\right\|}=\frac{\left(\left[R \boldsymbol{x}_i\right]_{\times} \boldsymbol{x}_i^{\prime}\right)^T\left[\boldsymbol{x}_i^{\prime}\right]_{\times}}{\theta_i^2} \boldsymbol{t} \triangleq \frac{1}{\theta_i^2} \boldsymbol{h}_i^T \boldsymbol{t}, \\
& \frac{\left\|\boldsymbol{t} \times R \boldsymbol{x}_i\right\|}{\left\|\boldsymbol{x}_i^{\prime} \times R \boldsymbol{x}_i\right\|}=\frac{\left(\left[R \boldsymbol{x}_i\right]_{\times} \boldsymbol{x}_i^{\prime}\right)^T\left[R \boldsymbol{x}_i\right]_{\times}}{\theta_i^2} \boldsymbol{t} \triangleq \frac{1}{\theta_i^2} \boldsymbol{h}_i^{\prime T} \boldsymbol{t},
\end{aligned}
\end{equation}
where $\theta_i=\left\|\boldsymbol{x}_i^{\prime} \times R \boldsymbol{x}_i\right\|$. Substituting into \cref{eq:eq7} and left-multiplying $[\boldsymbol{x}_i^{\prime}]_{\times}$, the LiGT constraint regarding the relative two-view translation is derived in~\cite{cai2021pose} as:
\begin{equation}
\left([\boldsymbol{x}_i^{\prime}]_{\times} R \boldsymbol{x}_i \boldsymbol{h}_i^T+\theta_i^2 [\boldsymbol{x}_i^{\prime}]_{\times}\right) \boldsymbol{t} \triangleq L_i(R,\boldsymbol{x}_i,\boldsymbol{x}_i^{\prime}) \boldsymbol{t}=0
\label{eq:eq8}
\end{equation}
Here, a two-view LiGT optimization (LiGTopt) is given by 
\begin{equation}
argmin_{(R, \boldsymbol{t})}\sum_i{||\tilde L_i \boldsymbol{t}||},
\label{}
\end{equation}
where $\tilde L_i \triangleq L(R,\tilde{\boldsymbol{x}}_i,\tilde{\boldsymbol{x}}_i^{\prime})$. As seen from the derivation process from \cref{eq:eq4} to \cref{eq:eq5}, both the PPO constraint and the above LiGT constraint encompass the coplanar essential matrix equation and the Kneip constraint. With these imaging constraints, two pose-only residuals can be formed, i.e., 
\begin{equation}
\begin{aligned}
v_i^{LiGT}(R, \boldsymbol t,\tilde{ \boldsymbol x}_i,\tilde{\boldsymbol x}^{\prime}_i) & =||\tilde L_i \boldsymbol{t}||, \\
v_i^{P P O}(R, \boldsymbol t, \tilde{\boldsymbol x}_i,\tilde{\boldsymbol x}^{\prime}_i) & =||\tilde{\boldsymbol \epsilon}_i - \tilde{\boldsymbol x}_i^{\prime}||,
\label{eq:v_LiGT}
\end{aligned}
\end{equation}
where ${\boldsymbol \epsilon}_i=\left\|\boldsymbol{t} \times \tilde{\boldsymbol x}_i^{\prime}\right\| R \tilde{\boldsymbol x}_i+\left\|\tilde{\boldsymbol x}_i^{\prime} \times R \tilde{\boldsymbol x}_i\right\| \boldsymbol{t}$ .

\section{ Six-Point Algorithm for Linear Relative Pose (LiRP)}
\label{sec:LiRP}
Let \( Q \) represent the matrix to be solved in the essential equation \cref{eq:eq4}, we have
\begin{equation}
 \boldsymbol{x}_i^{\prime T} Q \boldsymbol{x}_i = 0 \
\Leftrightarrow (\boldsymbol{x}_i^T \otimes \boldsymbol{x}_i^{\prime T})\boldsymbol q = 0 \stackrel{\triangle}{=} A_i \boldsymbol q=0,
\end{equation}
where \( \boldsymbol q \) denotes the vectorized form of the matrix \( Q \) arranged column-wisely, and $A_i$ is the i-th row vector of $A$ matrix.  Assume $\sigma_{1} \geq \sigma_{2} \geq \sigma_{3}$ are the three smallest singular values of matrix $A$, with their corresponding singular vectors \(\boldsymbol q_1\), \(\boldsymbol q_2\), and \(\boldsymbol q_3\). As $rank(A) = 6$ in planar scene or pure rotational motion, the true solution of  \( \boldsymbol q \) will be in the three-dimension solution space spanned by \(\boldsymbol q_1\), \(\boldsymbol q_2\), and \(\boldsymbol q_3\), i.e.,
\begin{equation}
\boldsymbol q = a \boldsymbol q_1 + b \boldsymbol q_2 + c \boldsymbol q_3,
\end{equation}
where $a$, $b$, and $c$ are coefficients to be determined. The solution $\boldsymbol q$ is determined up to a scale. Subsequently, the determination of these coefficients can be discussed by two cases:

Case (1): Let $c = 1$, then the values of $a$ and $b$ can be determined such that \( Q \) is an essential matrix satisfying the Demazure constraint~\cite{faugeras1993three}:
\begin{equation}
Q Q^{T} Q - \frac{1}{2} \text{trace}(Q Q^{T}) Q = 0
\Leftrightarrow B \boldsymbol y=0,
\label{eq:By=0}
\end{equation}
where $B$ is a 9x10 matrix constructed from \(\boldsymbol q_1\), \(\boldsymbol q_2\), and \(\boldsymbol q_3\), and $\boldsymbol y =(a^3, a^2b, ab^2, b^3, a^2, ab, b^2, a, b, 1)^T$. Assuming that the first four columns form a submatrix $B_1$ and the remaining columns constitute a submatrix $B_2$, that is, $B=(B_1|B_2)$. By left multiplying \cref{eq:By=0} with the pseudo-inverse of $B_1$, denoted as $B_1^+$, we obtain $(I_{4\times4} | M_{4\times6}) \boldsymbol y=0$.
The left-side structure of the above equation can be illustrated as in \cref{table1}, of which the first row lists the monomial elements of $\boldsymbol y$, the horizontal bars "-" represent specific values of \(M\), and other blank entries are 0s. Notably, each monomial of third order in $\boldsymbol y$ can be expressed as linear combinations of those of lower orders in the same row. In this regard, $(I | M) \boldsymbol y=0$ gives rise to the equations \(C_1\boldsymbol g = a\boldsymbol{g}\) and \(C_2\boldsymbol g = b\boldsymbol{g}\) of eigensystem form, where \(\boldsymbol g = (a^2, ab, b^2, a, b, 1)^T\). Therefore, by determining the eigenvectors of $6$-by-$6$ action matrices \(C_1\) and \(C_2\), as many as 12 potential solution vectors for \(\boldsymbol g\) can be derived. The coefficients \(a\) and \(b\) can be directly determined from the last three elements of each solution vector, specifically, \(a = g_4/g_6\) and \(b = g_5/g_6\). In this scenario, we obtain 12 candidate solutions for $\boldsymbol q$, denoted as $\boldsymbol q_s$.

\begin{table}[h]
    \centering
    \begin{tabular}{|c|c|c|c|c|c|c|c|c|c|}
        \hline
        $a^3$ & $a^2b$ & $ab^2$ & $b^3$ & $a^2$ & $ab$ & $b^2$ & $a$ & $b$ & $1$ \\
        \hline
        1 & & & & - & - & - & - & - & - \\
        \hline
        & 1 & & & - & - & - & - & - & - \\
        \hline
        & & 1 & & - & - & - & - & - & - \\
        \hline
        & & & 1 & - & - & - & - & - & - \\
        \hline
    \end{tabular}
    \caption{Polynomial system.}
    \label{table1}
\end{table}

Case (2). If $c = 0$, then we deduce $\boldsymbol q = a \boldsymbol q_1 + b \boldsymbol q_2$. Analogously, setting $b = 1$ and  only $a$ needs to be determined. Considering the property that the determinant of the essential matrix is zero, a polynomial in terms of \(a\) can be constructed by
\begin{equation}
det(Q)=0 \Leftrightarrow \boldsymbol d^T \boldsymbol z=0,
\label{eq:Dz=0}
\end{equation}
where  \(\boldsymbol z = (a^3, a^2, a, 1)^T\) . In analogy to Case (1), three candidate solutions for $\boldsymbol q$ can be derived, say $\boldsymbol q^\prime_s$. Setting $b = 0$, $\boldsymbol q_1$ also emerges as a potential solution. Taking into account different solution sequences, $\boldsymbol q_2$ and $\boldsymbol q_3$ can similarly be considered as candidate solutions.

In summary, the above analysis yields 18 candidate solutions for $\hat{\boldsymbol q} = (\boldsymbol q_s,\boldsymbol q^\prime_s, \boldsymbol q_1,\boldsymbol q_2,\boldsymbol q_3 )$. Each candidate solution can be decomposed into four relative pose solutions~\cite{hartley2003multiple}. With the aid of the relative pose identification's inequality strategy revealed in~\cite{cai2019equivalent}, that is to say, $M_1(R)>0$ and $M_2(R,\boldsymbol t)>0$ therein, eighteen candidate solutions of the essential matrix can be processed to yield the same number of relative pose candidate solutions. Subsequently, the identification will further rely on minimizing $\boldsymbol v^{PPO}$, to be explained in next section.

\begin{algorithm}[htbp]
\caption{Weighted LiRP Algorithm}\label{alg:LiRP algorithm}
\begin{algorithmic}[1]
\Require
  point pairs $\left\{ (\tilde{\boldsymbol x}_1, \tilde{\boldsymbol x}^{\prime}_1),...,(\tilde{\boldsymbol x}_n, \tilde{\boldsymbol x}^{\prime}_n) \right\}$,
  (optional) weights $\boldsymbol w$ 
\Ensure
relative rotations $R$ and translations $\boldsymbol t$
\State construct weighted $A$ matrix $\gets $ $A_i = w_i(\tilde{\boldsymbol x}_i^T \otimes \tilde{\boldsymbol x}_i^{\prime T})$
\State obtain $\boldsymbol q_1,\boldsymbol q_2,\boldsymbol q_3 \gets $ SVD decomposition of A matrix
\State compute $B$ matrix by \cref{eq:By=0}
\State obtain $M$ matrix by left multiplying $B_1^+B$
\State construct $C_1$, $C_2$ $\gets M$
\State $\boldsymbol q_s \gets $ eigenvectors of $C_1$, $C_2$
\State $\boldsymbol q^\prime_s \gets $ solve polynomial by \cref{eq:Dz=0}
\State $\hat{\boldsymbol q} \gets$ $\boldsymbol q_s,\boldsymbol q^\prime_s, \boldsymbol q_1,\boldsymbol q_2,\boldsymbol q_3 $
\State $R, \boldsymbol t \gets $$\hat{\boldsymbol q}$
\end{algorithmic}
\end{algorithm}
\section{Brief Summary of GNC-RANSAC method }
\label{sec:GNC-RANSAC method}


The recent GNC-base IRLS method~\cite{peng2023convergence} alternates between optimizing a smooth majorizer function and increasing a parameter $\mu$ at each iteration, which is designed to accelerate convergence and yield stable outcomes. The majorizer function, denoted as $q(v_i, \mu)$, represents a shifted version of the objective function $\rho(v_i)$, which is to be minimized. The parameter $\mu$ controls the tradeoff between accuracy and robustness. By alternately updating the majorizer function and increasing $\mu$, GNC~\cite{blake1987visual} aims to find a stationary point of the objective function to reach convergence. In general IRLS problems, the scale-invariant issue is often considered. Let $\sigma$ denote the standard deviation of the sample. In this paper, based on the recommendation of Huber~\cite{huber2004robust}, we use the median of absolute deviations to determine $\sigma$, i.e., $\sigma=1.4826 \times \text{med}|\hat{v}_i-\text{med}(\hat{v}_i)|$.
By integrating the GNC-IRLS with the LiRP algorithm, we developed a robust relative pose estimation method, as shown in \cref{alg:IRLS-Pose-only}. For instance, with 30 point pairs, it can resist nearly 40\% outlier fraction. However, for higher outlier fractions, the RANSAC scheme is still needed to increase robustness, as seen in \cref{alg:Gnc-ransac}. Compared to the standard RANSAC method, our robust relative pose estimation algorithm for fitting model relaxes the requirement that all sub-samples must consist entirely of inliers. Instead, it ensures that the outlier fraction in sub-samples does not exceed a certain threshold, such as 40\%. This approach makes it easier to find suitable sub-samples at higher outlier fractions.

\begin{algorithm}[htbp]
\caption{GNC-IRLS Pose Estimation Algorithm}\label{alg:IRLS-Pose-only}
\begin{algorithmic}[1]
\Require
  Point pairs $\left\{ (\tilde{\boldsymbol x}_1, \tilde{\boldsymbol x}^{\prime}_1), \ldots, 
  (\tilde{\boldsymbol x}_n, \tilde{\boldsymbol x}^{\prime}_n) \right\}$, stop threshold $\epsilon_0$, 
  maximum iteration $n_{iter}$
\Ensure
  The model parameters $R, \boldsymbol t$

\State Initialize $\epsilon^{(0)}$, $\mu^{0}$, and weights $\boldsymbol w^{(0)}$,
\State $converged \gets \text{false}$

\vspace{0.1cm}
\For{$k \gets 0$ to $n_{iter}-1$}
    \State Obtain $R^{(k)}$, $\boldsymbol t^{(k)}$ with $\boldsymbol w^{(k)}$ by \cref{alg:LiRP algorithm}
    \State Compute residual vector $\boldsymbol{v}^{LiGT}$ by \cref{eq:v_LiGT} 
    \State $\sigma^{(k)} \gets 1.4826 \cdot \text{median}(|\boldsymbol v^{LiGT}|)$
    \State $c^{(k)} \gets 5.54 \cdot \sigma^{(k)}$
    \State $\boldsymbol w^{(k+1)} \gets$ Update weights by $\boldsymbol{v}^{LiGT}$, 
           $c^{(k)}$, and $\mu^{(k)}$ according to \cite{peng2023convergence}
    \State $\epsilon^{(k+1)} \gets \sum w_i^{(k+1)}{v}_i^{LiGT}$
\vspace{0.1cm}
    \If{$|\epsilon^{(k+1)} - \epsilon^{(k)}| < \epsilon_0$} 
        \State $converged  \gets \text{true}$
        \State \textbf{exit loop} 
    \EndIf
    \State Update $\mu^{(k+1)}$ by super-linear schedule \cite{peng2023convergence}
\EndFor
\vspace{0.1cm}
\end{algorithmic}
\end{algorithm}

\begin{algorithm}[htbp]
\caption{GNC-RANSAC Algorithm}\label{alg:Gnc-ransac}
\begin{algorithmic}[1]
\Require
  point pairs $P=\left\{ (\tilde{\boldsymbol x}_1, \tilde{\boldsymbol x}^{\prime}_1),...,(\tilde{\boldsymbol x}_n, \tilde{\boldsymbol x}^{\prime}_n) \right\}$, stop threshold $\epsilon_0$, sample size $n_{s}$, $n_{iter}$, inlier threhold $\theta$
\Ensure
the model parameters $R, \boldsymbol t$

\State $n_{maxinliers} \gets 0$
; $R_{best}, \boldsymbol t_{best},\gets \emptyset$ 
\For{$k \gets 0$ to $n_{iter}-1$}
\State $ P_s \gets$ randomly select $n_s$ sample from $P$
\State $R_s, \boldsymbol t_s \gets$ fit model using \cref{alg:IRLS-Pose-only} with $P_s$
\State $\boldsymbol v^{L i G T} \gets$ obtain residual vector by \cref{eq:v_LiGT} for $P$
\State $P_{inliers} \gets$ $\boldsymbol v^{L i G T}$, $\theta$

\If{$|P_{inliers}| > n_{maxinliers}$}

\State $n_{maxinliers} \gets |P_{inliers}|$
\State $R_{best}, \boldsymbol t_{best} \gets R_s, \boldsymbol t_s$
\State $P_{best} \gets P_{inliers}$
\EndIf
\EndFor
\State $R, \boldsymbol t \gets$ fit model using \cref{alg:IRLS-Pose-only} with $P_{best}$

\end{algorithmic}
\end{algorithm}
\section{Experiments}
\label{sec:Experiments}

\begin{figure}
    \centering
    \begin{subfigure}{0.5\linewidth}
        \includegraphics[width=\linewidth]{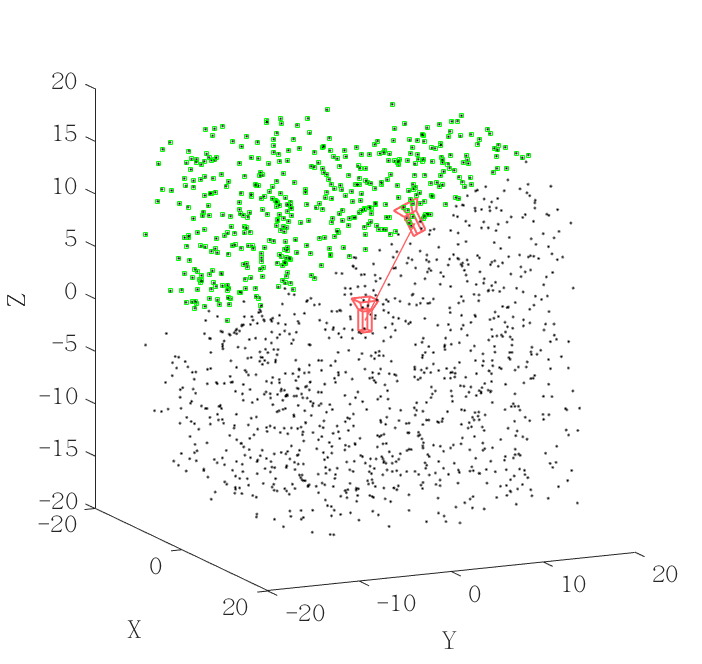}
        \caption{Normal scenario}
        \label{fig:normal RTX}
    \end{subfigure}%
    \begin{subfigure}{0.5\linewidth}
        \includegraphics[width=\linewidth]{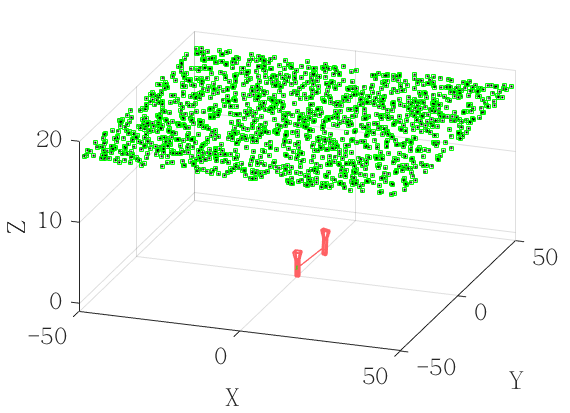}
        \caption{Planar scenario}
        \label{fig:planar RTX}
    \end{subfigure}
    \caption{Simulation of two-view structure and 3D feature points. The green nodes represent 3D points that conform to chirality constraints, and the black nodes represent 3D points otherwise. The two red cameras represent the pose of the two views. The red line indicates the displacement baseline.}
    \label{fig:RTX}
\end{figure}

The global translation has been well solved by the LiGT algorithm that is linear in nature and capable to produce nearly optimal global translation if provided excellent input of rotations~\cite{cai2021pose}. Unlike traditional methods~\cite{pizarro2003relative,nister2004efficient,stewenius2006recent,zhao2020efficient} that have mostly emphasized the accuracy and robustness of relative translation, the current work focuses on the performance of relative rotation estimation. 

In both simulation and real testing, the sampling size for GNC-RANSAC is set to $n_{s}=30$ and the maximum iteration of RANSAC is consistently set to 50 for all algorithms under investigation.

\subsection{Synthetic Experiment} 
The synthetic scenarios are of two types: 1) a normal scenario generated by using the OpenGV's code to construct 3D feature points and camera poses of two views, retaining only those bearing vectors satisfying chirality constraints (see \cref{fig:RTX}); 2) a planar scenario intentionally generated to examine the algorithm's behavior in a planar scene. The synthetic experiment was conducted in the Ubuntu 18.04 environment running on a single core. Monte Carlo tests were repeated 500 times. The image matching outliers were generated by a predetermined outlier fraction.  Before running the experiment, we categorized the compared algorithms into two groups: the first group includes initial relative pose estimation methods, including the Pizarro's 6pt algorithm~\cite{pizarro2003relative} , the LiRP algorithm, and 5pt~\cite{stewenius2006recent} / 7pt~\cite{hartley2003multiple}  / 8pt~\cite{hartley2003multiple} algorithms supported by OpenGV. The second group comprises optimization techniques, including the $N$-point method  (Npt)~\cite{zhao2020efficient} , eigensolver~\cite{kneip2013direct} / nonlinear optimization~\cite{kneip2014opengv} in OpenGV, BA~\cite{hartley2003multiple}, and the LiGT optimization.


\subsubsection{Robustness of relative pose estimation} 

%


\begin{table*}[!htbp]
\centering
\caption{Structure degenerate problem for relative rotation estimation. The pair (x,y) represents 'x' as the mean rotation error and 'y' as the minimum distance $d_{\min}$ between the eigenvalues of the action matrix. The symbol ‘-’ signifies that no data is provided. The labels (5) and (n) in column headers correspond to results estimated using five and $n$ point pairs, respectively. The multiple solutions of essential matrix herein are identified using the true essential matrix.}
\label{tab:E_mat_problem}
\begin{tabular}{|c|c|c|c|c|c|}
\hline
Scenes & Noise Pixel & 5pt (5)~\cite{stewenius2006recent}& 5pt  (m)~\cite{stewenius2006recent}& LiRP (m) & Npt (m)~\cite{zhao2020efficient} \\
\hline
\multirow{2}{*}{Planar} & 0    & $(0, 4.19e^{-1})$ & $(2.79e^{-2}, 1.08e^{-15})$ & $(0, 1.44)$ & $(3.48e^{-2}, -)$ \\
\cline{2-6}
                        & 0.1  & $(3.36e^{-3}, 4.45e^{-1})$  & $(2.75e^{-4}, 1.40e^{-4})$ & $(\boldsymbol{2.54e^{-4}}, 1.28)$ & $(4.38e^{-2}, -)$ \\
\hline
\multirow{2}{*}{Normal} & 0    & $(0, 2.27e^{-1})$ & $(0, 7.80e^{-3})$ & $(0, 1.23)$ & $(0, -)$ \\
\cline{2-6}
                        & 0.1  & $(2.58e^{-3}, 3.10e^{-1})$ & $(5.47e^{-4}, 7.56e^{-3})$ & $(\boldsymbol{2.03e^{-4}}, 1.32)$ & $(2.16e^{-4}, -)$ \\
\hline
\end{tabular}
\end{table*}

Unlike the RANSAC method, IRLS typically requires reweighting almost all matches. The Stewenius method, a well-established approach for estimating relative pose, is traditionally employed for sets of five matching points and can also be extended to handle more than five pairs of points ($n\geq6$)~\cite{stewenius2006recent}. It is widely recognized that both the 7pt and 8pt methods exhibit planar degeneracy. It has been pointed out~\cite{stewenius2006recent,pizarro2003relative} that the linear 6-point algorithm~\cite{philip1996non} also suffers from planar degeneracy, and the Pizarro's 6pt method has been found to be prone to instability in solutions.

The noise-free experiment in \cref{tab:E_mat_problem} throws lights on  theoretical limitation of the Stewenius method when addressing $n$-point pairs ($n=30$ herein) within planar configurations. Specifically, under the planar scene with a noise-free condition, the action matrix constructed by using the Stewenius method~\cite{stewenius2006recent} for $n$ point pairs has repeated eigenvalues. The multiple eigenvectors corresponding to the repeated eigenvalues lead to infinite solutions of essential matrix. We utilized an indicator $ d_{\min} = \min_{i,j} \{ |\lambda_i - \lambda_j| \}$ to quantify the minimum distance between the eigenvalues of the action matrix. In the case of a noise-free planar scene, the Stewenius method's $d_{\min}$ is zero, signifying the presence of duplicate eigenvalues. The rotation accuracy of $2.79e^{-2}$ indicates the above-mentioned limitation. Fortunately, when point pairs are subject to 0.1 pixel noise, this limitation is largely mitigated. The result in \cref{tab:E_mat_problem} also reveals the planar degenerate problem for the $N$-point method~\cite{zhao2020efficient}. In contrast, the LiRP algorithm consistently achieved the best accuracy in both normal and planar scenes, maintaining $d_{\min}$ above acceptable thresholds. The LiRP algorithm's rotation accuracy further confirms the absence of planar case limitations.

\begin{figure}
    \centering
    \includegraphics[width=0.25\textwidth]{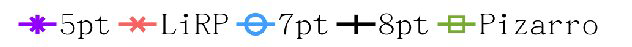}
 
    \begin{subfigure}{0.5\linewidth}
        \includegraphics[width=\linewidth]{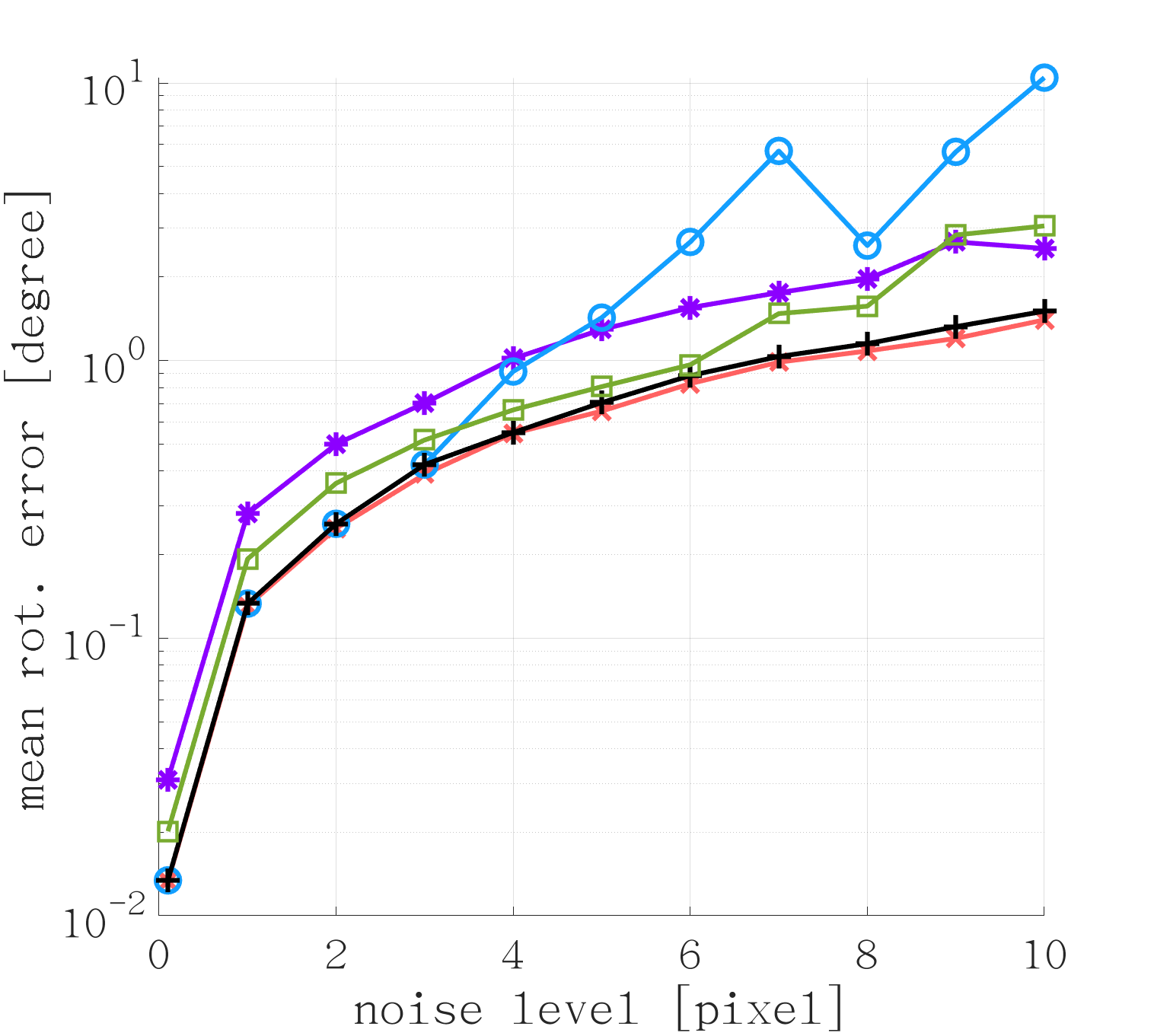}
        \caption{Normal scenario}
        \label{fig:noise_test_a}
    \end{subfigure}%
    \begin{subfigure}{0.5\linewidth}
        \includegraphics[width=\linewidth]{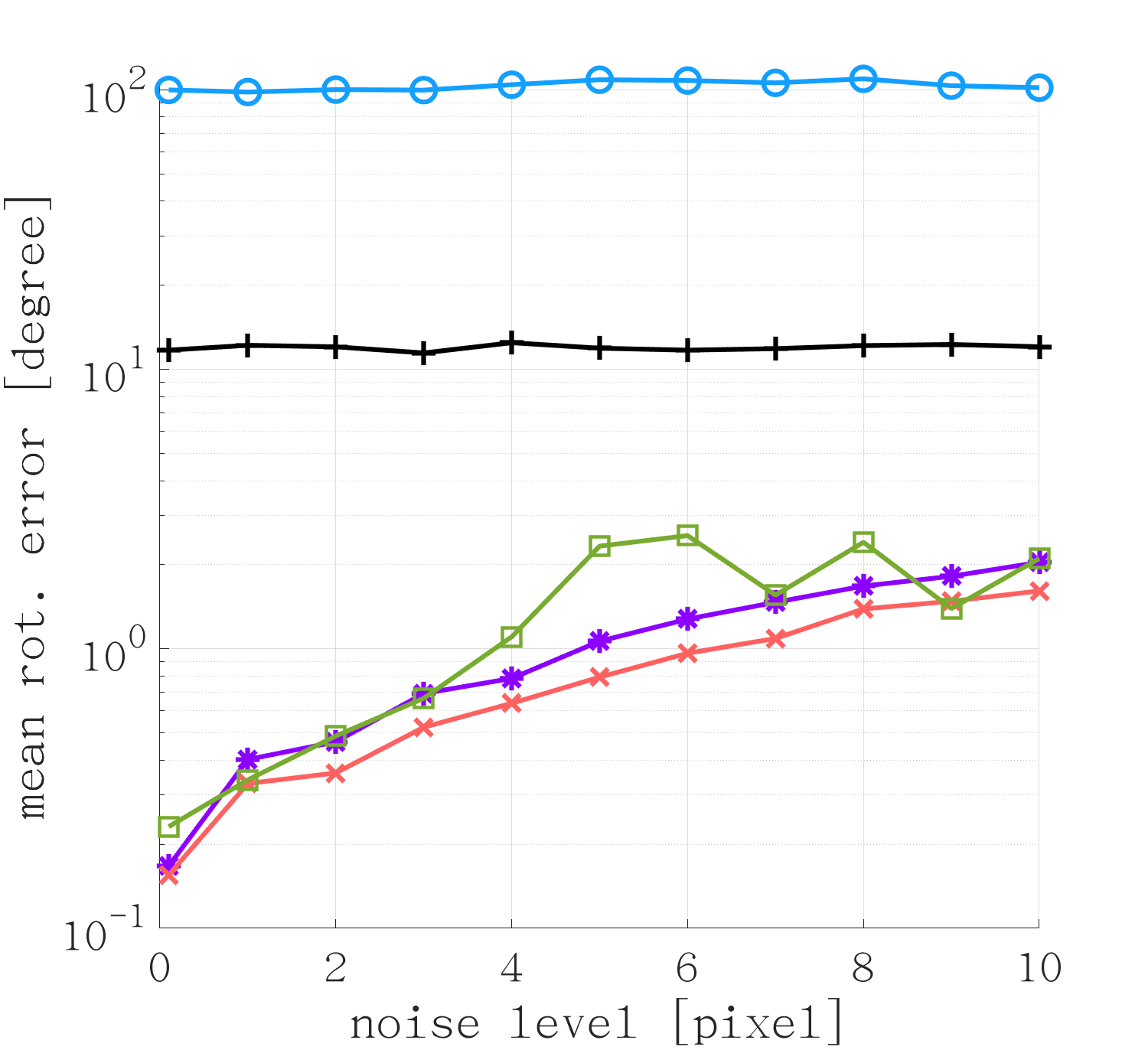}
        \caption{Planar scenario}
        \label{fig:noise_test_b}
    \end{subfigure}
    \caption{Noise test results of initial relative pose estimation methods.}
    \label{fig:test_ini_methods}
\end{figure}

\begin{figure}
    \centering
    \includegraphics[width=0.25\textwidth]{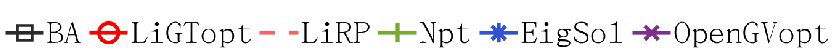}
 
    \begin{subfigure}{0.5\linewidth}
        \includegraphics[width=\linewidth]{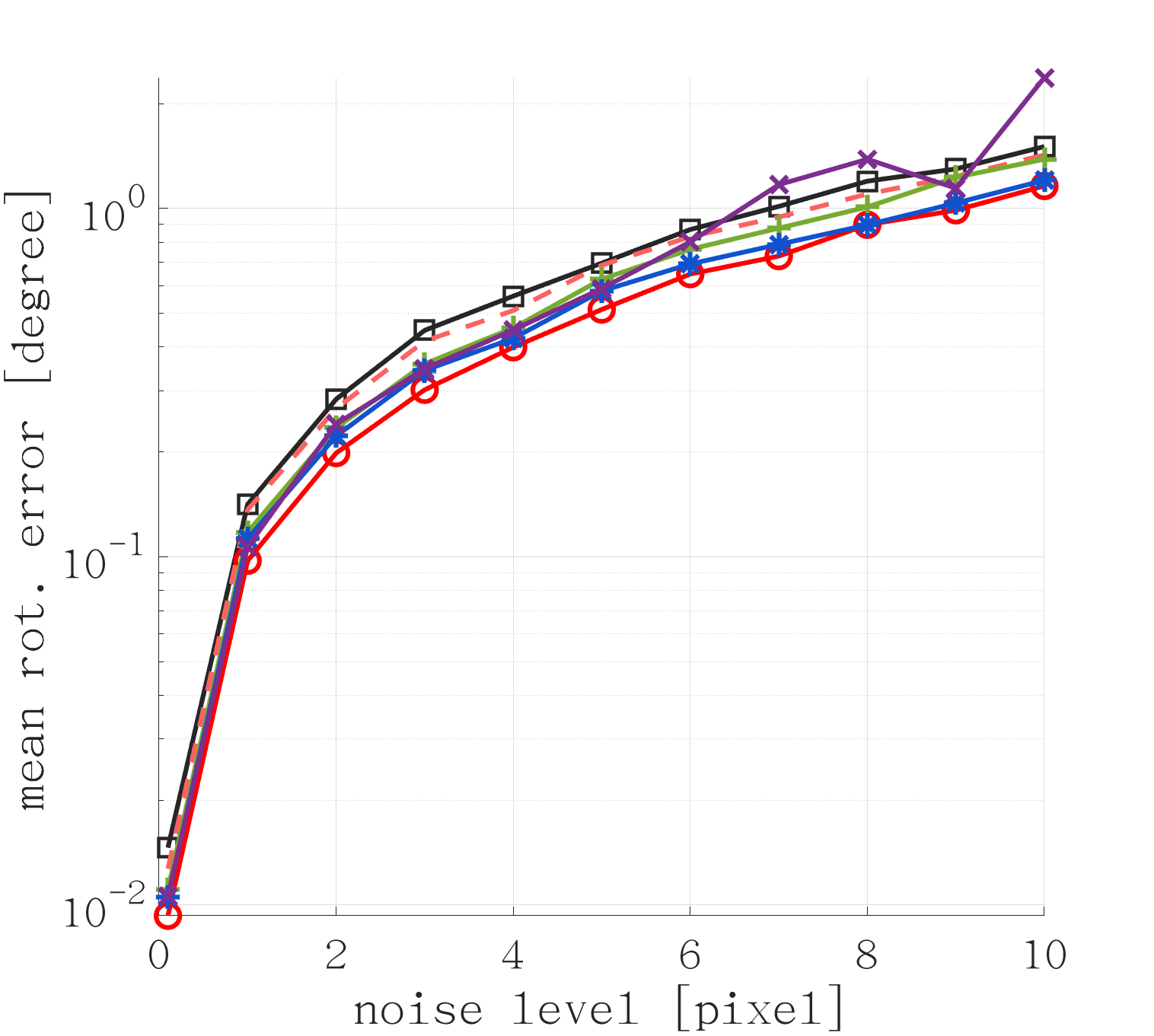}
        \caption{Normal scenario}
        \label{fig:noise_opt_test_a}
    \end{subfigure}%
    \begin{subfigure}{0.5\linewidth}
        \includegraphics[width=\linewidth]{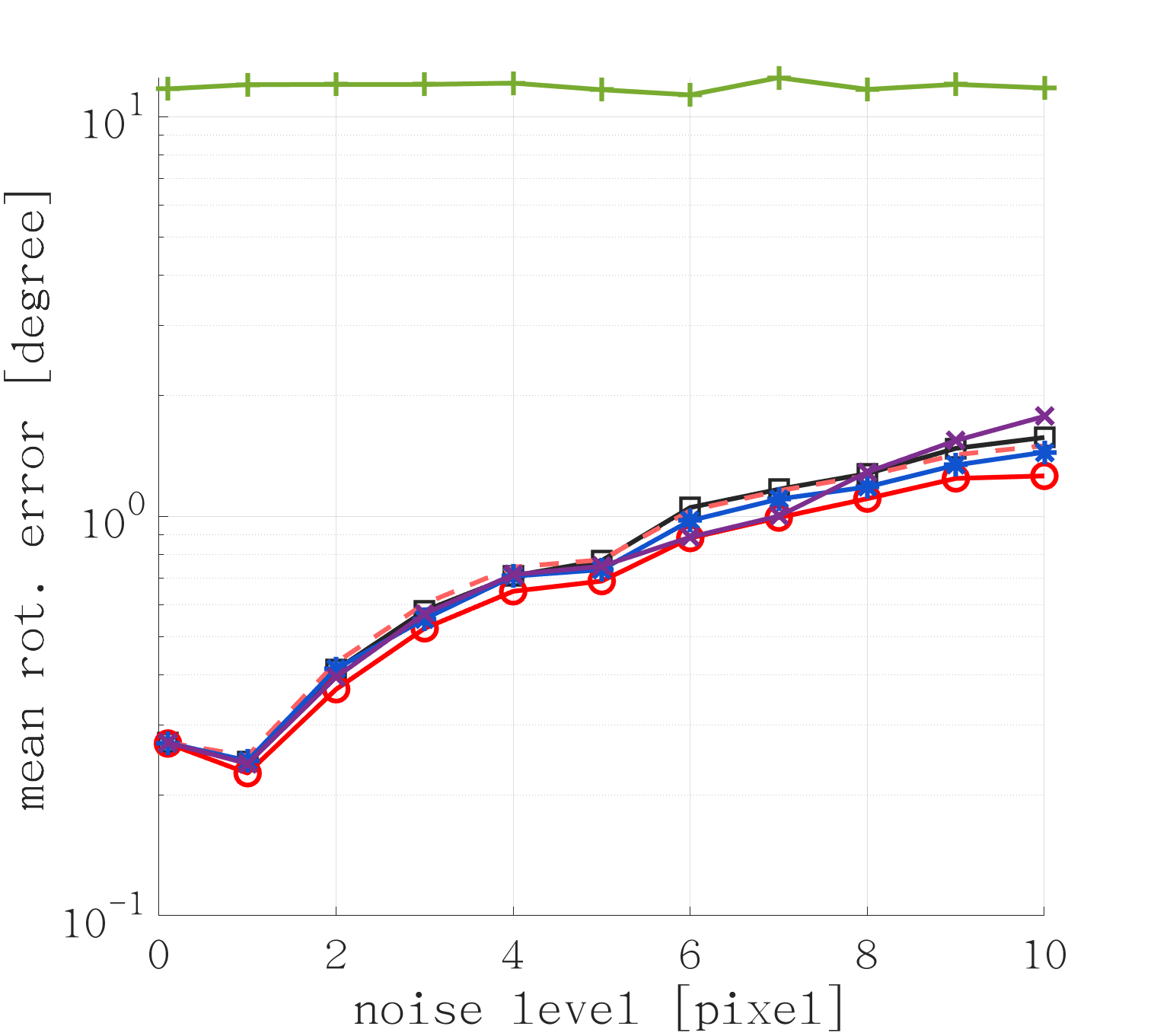}
        \caption{Planar scenario}
        \label{fig:noise_opt_test_b}
    \end{subfigure}
    \caption{Noise test results of optimization methods, which are all initialized by LiRP.}
    \label{fig:noise_opt_test}
\end{figure}

Subsequently, we conducted tests to evaluate the impact of noises on relative pose estimation methods for $n$ point pairs problem, see \cref{fig:test_ini_methods,fig:noise_opt_test}. We compare the performance without outliers across a spectrum of noise levels for 30 point pairs, which were incremented from 0 to 10 pixels in steps of one pixel.

In normal scenarios, as evidenced by \cref{fig:noise_test_a} and \cref{fig:noise_opt_test_a}, the LiRP estimation and the LiGT optimization both demonstrate outstanding accuracy. Notably, LiRP achieves the lowest mean rotational error, particularly as noise levels increase, in stark contrast to the 5pt and Pizarro's 6pt methods. Moreover, LiRP's rotational accuracy not only competes with but often surpasses that of BA optimization and is on par with the majority of other optimization techniques. Concurrently, LiGT optimization upholds the highest standard of rotational accuracy, even as noise levels escalate. Note that the Npt method are prone to degenerate problems in planar scenes, as shown in \cref{fig:noise_opt_test_b}. 

\subsubsection{Selection of Optimal Residual Statistics} 

\begin{figure}
    \centering
    \includegraphics[width=0.25\textwidth]{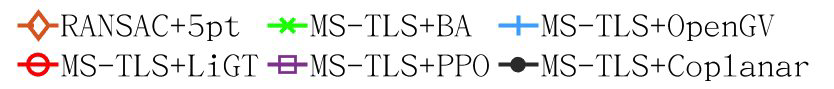}
 
    \begin{subfigure}{0.5\linewidth}
        \includegraphics[width=\linewidth]{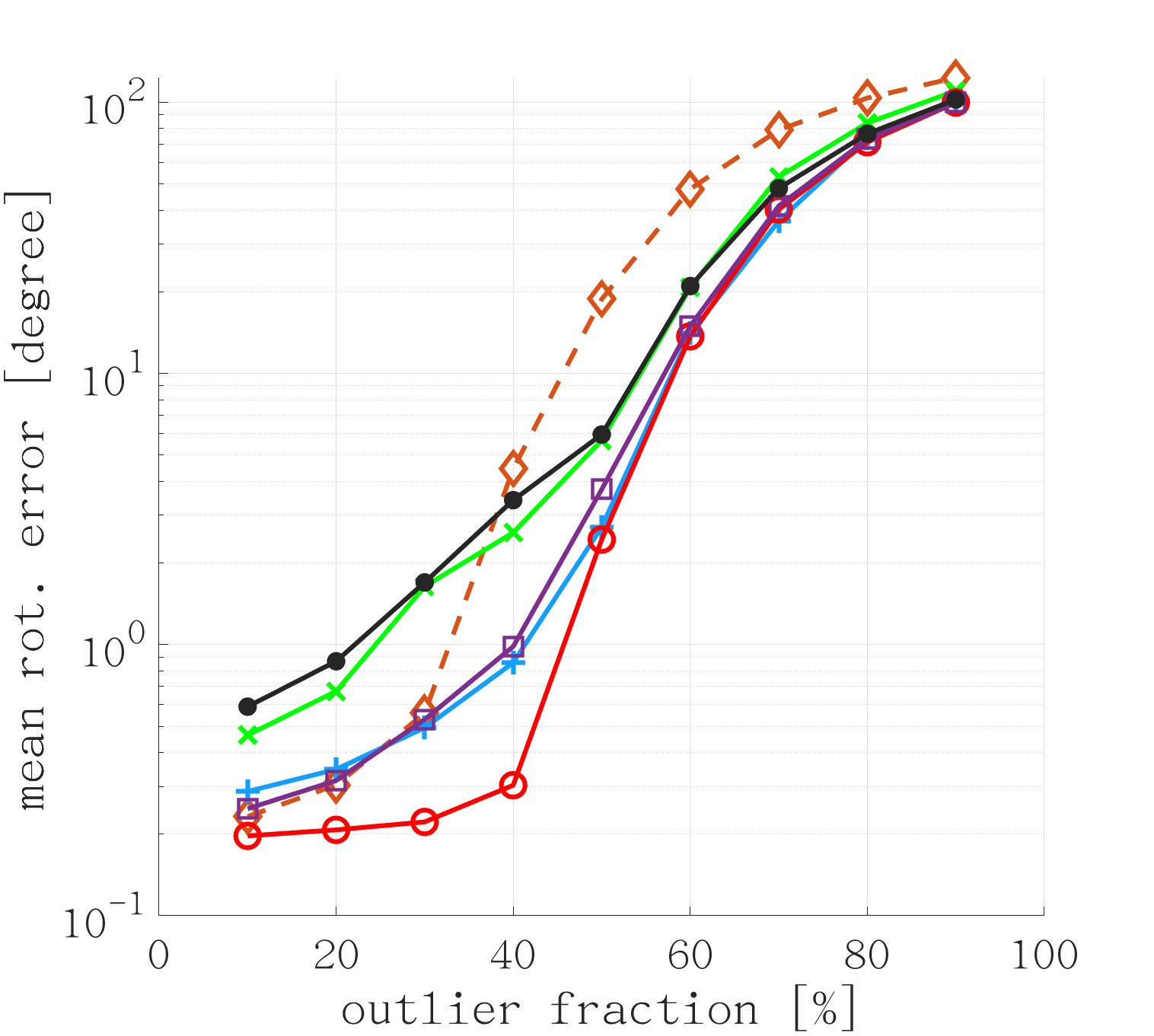}
        \caption{Normal scenario}
        \label{fig:outlier_v_test_a}
    \end{subfigure}%
    \begin{subfigure}{0.5\linewidth}
        \includegraphics[width=\linewidth]{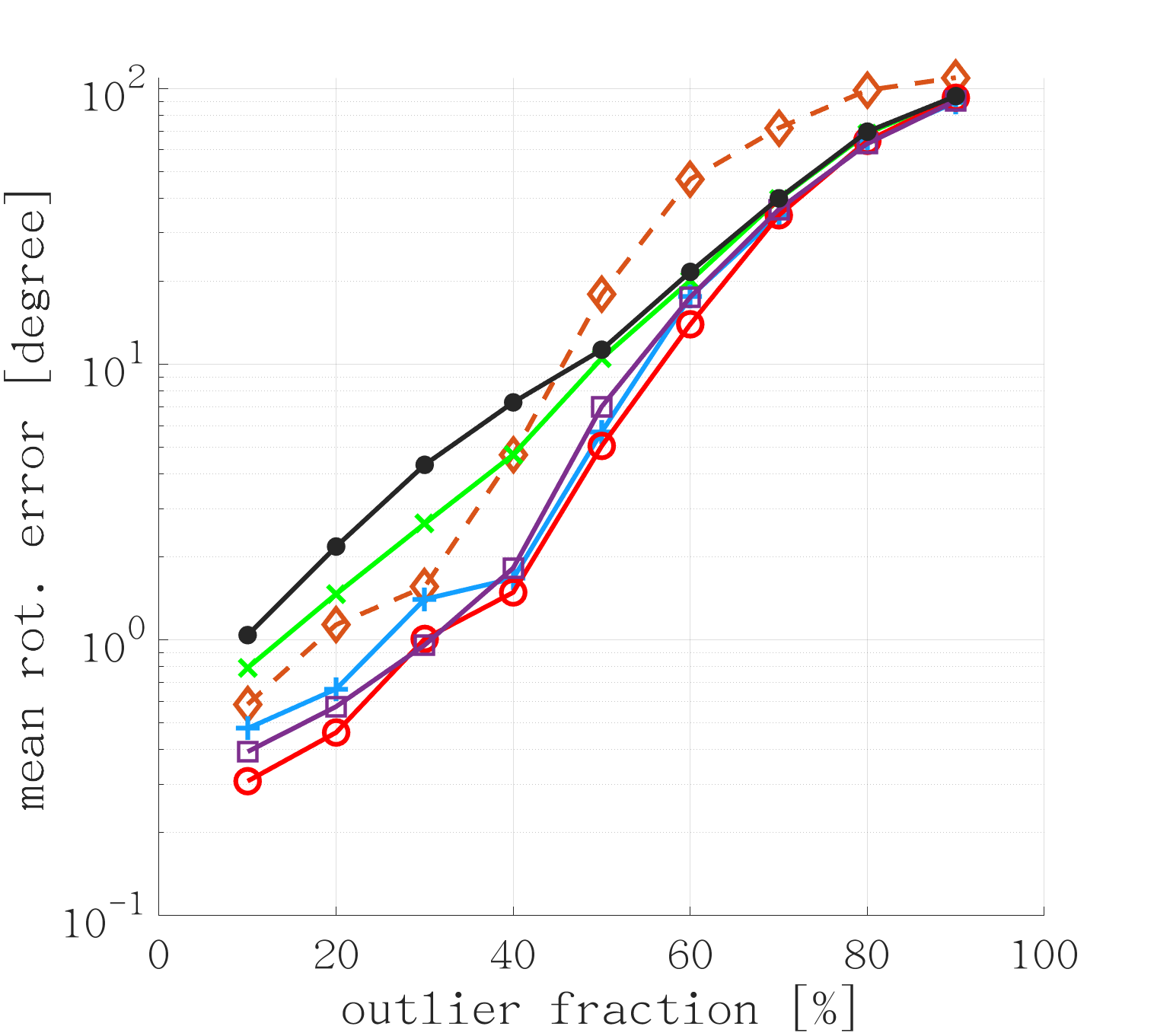}
        \caption{Planar scenario}
        \label{fig:outlier_v_test_b}
    \end{subfigure}
    \caption{Robustness test for different residuals. All variants of MS-TLS utilize LiRP for relative rotation. The dashed line represents the benchmark result (RANSAC+5pt) provided by OpenGV.}
    \label{fig:outlier_v_test}
\end{figure}

In the sequel, we compared the optional residual statistics in \cref{sec:Two-view Constraints and Residuals} and selected the optimal one to be combined with the GNC scheme in \cref{alg:IRLS-Pose-only}. With image noise fixed at 1 pixel, \cref{fig:outlier_v_test} displays the mean error results of different residual statistics across different outlier fraction ratios for $n = 30$. We use 'MS-TLS' to denote the majorization and superlinear GNC schedule for truncated least-squares loss~\cite{peng2023convergence}. In each subplot, we have incorporated the RANSAC-based 5pt method in OpenGV as comparison benchmark. 

From \cref{fig:outlier_v_test}, it is obvious that the MS-TLS stategy of utilizing $\boldsymbol v^{L i G T}$ as the re-weighted iterative residual statistic yields the highest rotation accuracy, which ensures that the rotation error remains less than $1^\circ$ for up to 40\% outlier fractions in normal scene (see \cref{fig:outlier_v_test_a}) and 30\% outlier fractions in planar scene (see \cref{fig:outlier_v_test_b}).

Considering the results from \cref{fig:outlier_v_test}, it is evident that using $\boldsymbol v^{LiGT}$ as the residual statistics for weighting is most appropriate in relative rotation estimation.

\subsubsection{Results of the GNC-RANSAC Scheme}

 

\begin{figure}
    \includegraphics[width=0.25\textwidth]{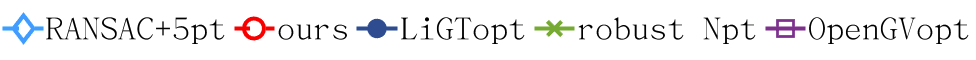}
    \centering
    \begin{subfigure}{0.5\linewidth}
        \centering
        \includegraphics[width=\linewidth]{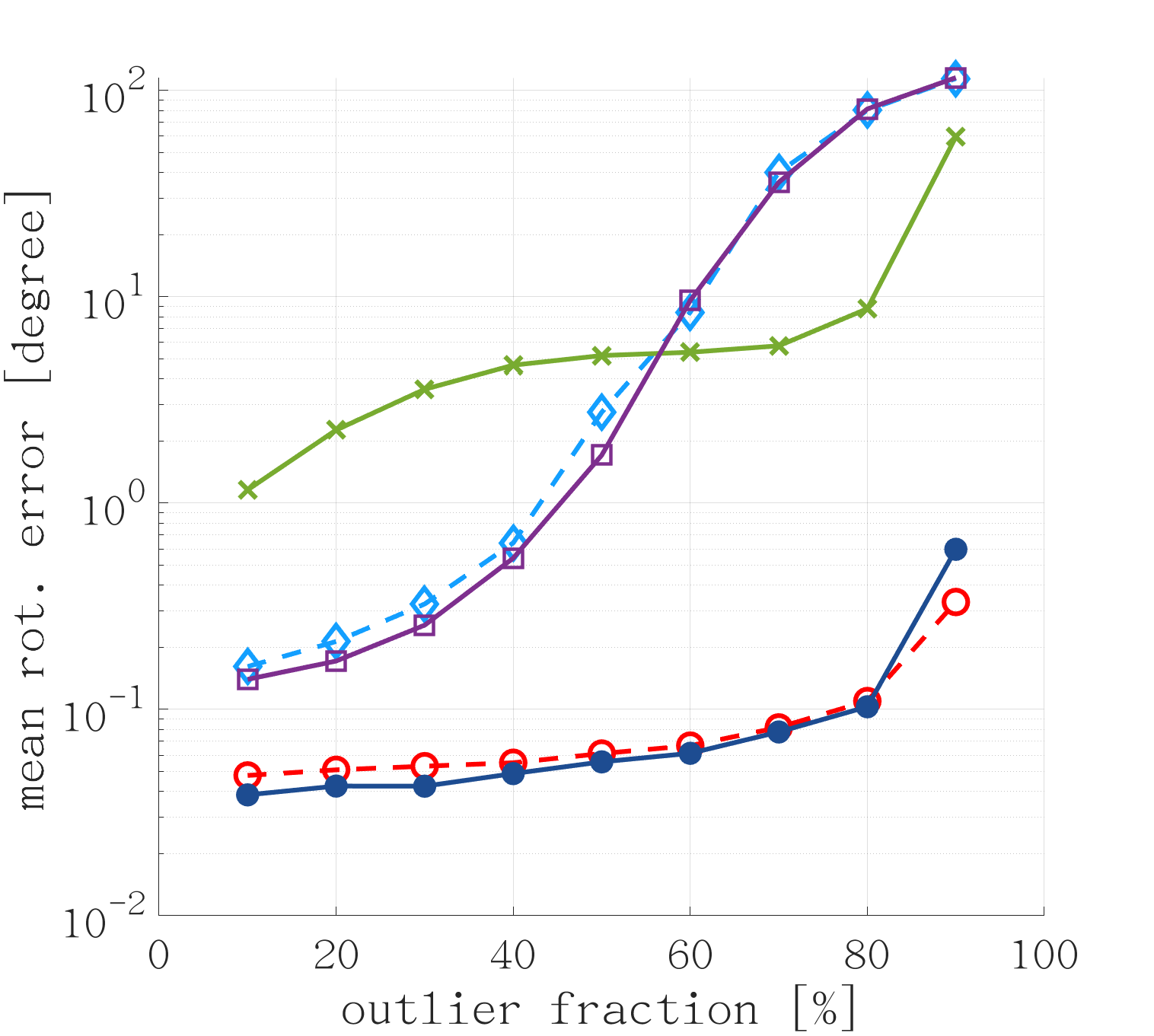}
        \captionsetup{justification=centering}
        \caption{}
        \label{fig:gnc_ransac_normal_mean}
    \end{subfigure}%
    \begin{subfigure}{0.5\linewidth}
        \centering
        \includegraphics[width=\linewidth]{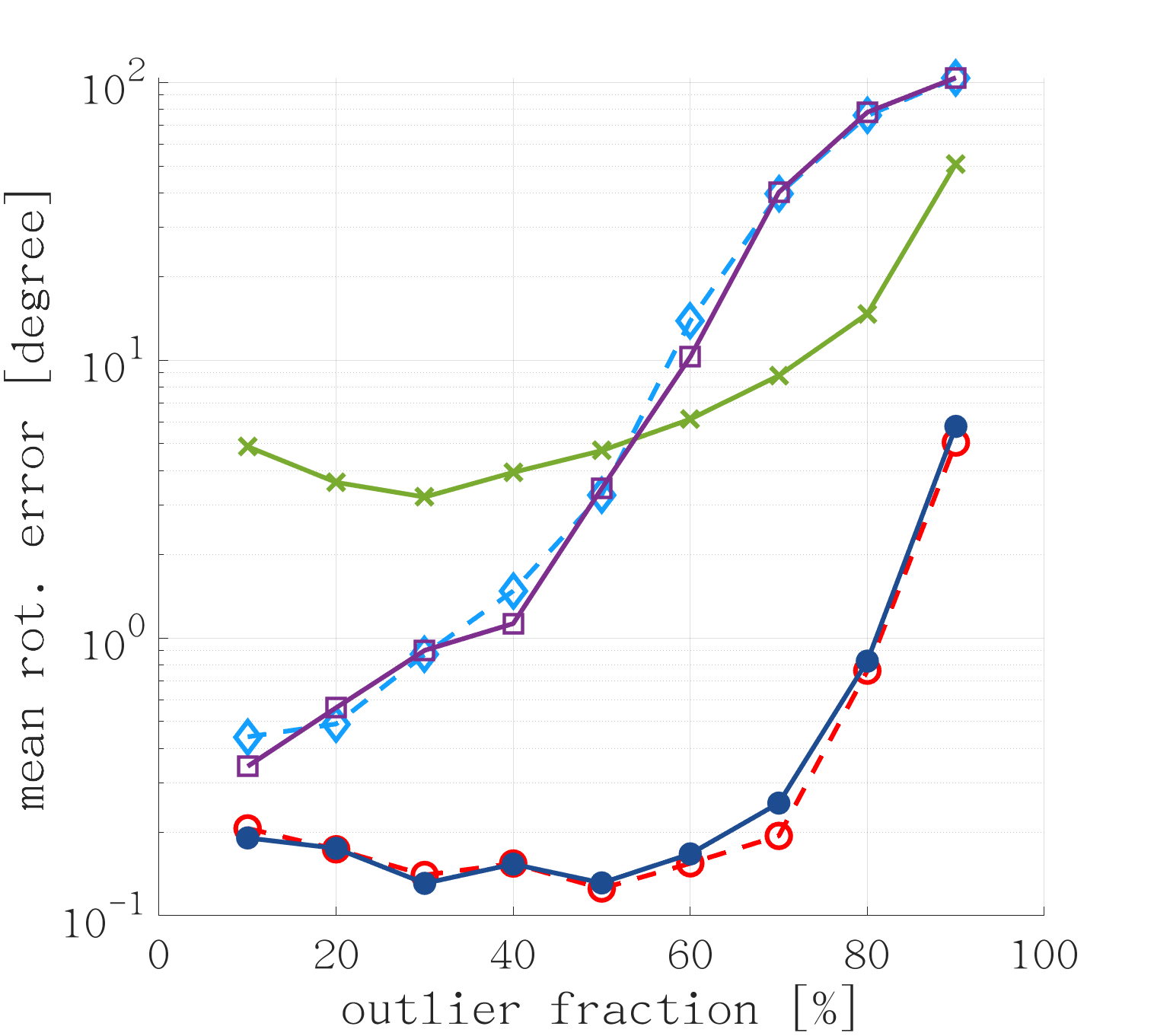}
        \captionsetup{justification=centering}
        \caption{}
        \label{fig:gnc_ransac_300c_planar_mean}
    \end{subfigure}
    
    \begin{subfigure}{0.5\linewidth}
        \centering
        \includegraphics[width=\linewidth]{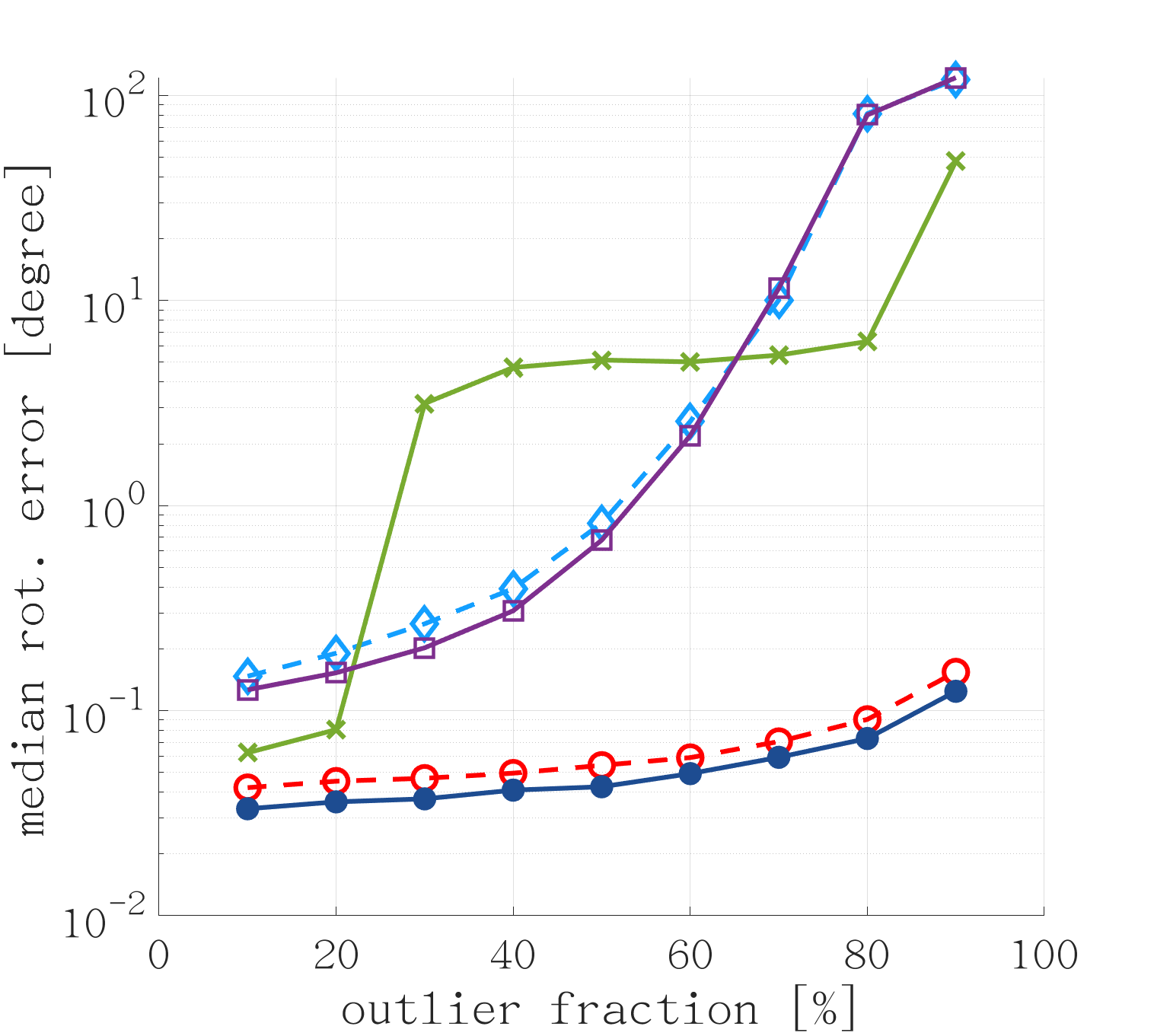}
        \captionsetup{justification=centering}
        \caption{}
        \label{fig:gnc_ransac_normal_median}
    \end{subfigure}%
    \begin{subfigure}{0.5\linewidth}
        \centering
        \includegraphics[width=\linewidth]{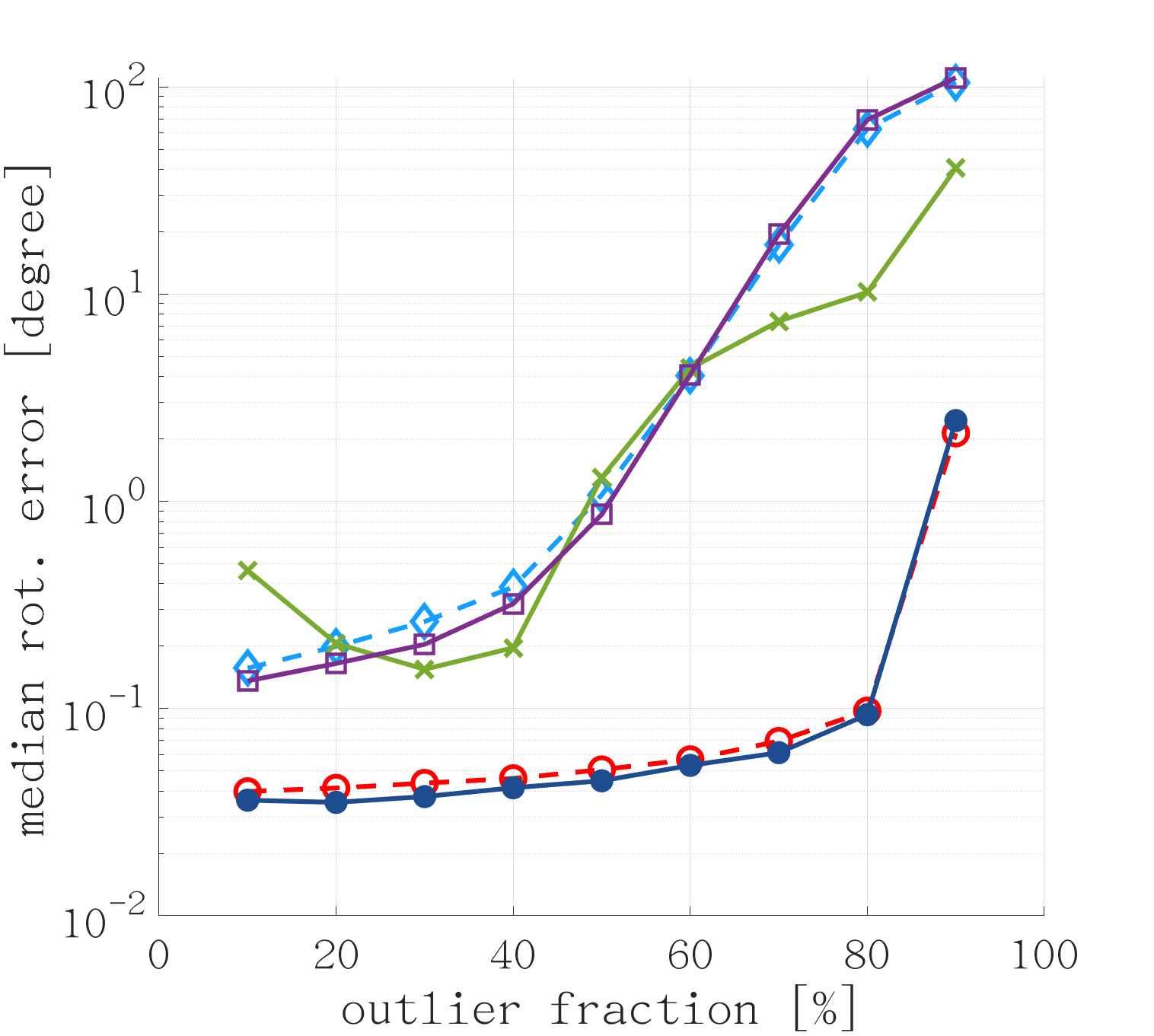}
        \captionsetup{justification=centering}
        \caption{}
        \label{fig:gnc_ransac_planar_median}
    \end{subfigure}
    \caption{Accuracy of the GNC-RANSAC fusion scheme as compared with state-of-art methods. LiGTopt is optimized based on our GNC-RANSAC scheme, and OpenGVopt is optimized based on OpenGV's RANSAC+5pt. The first and second columns are for normal scenes and planar scenes, respectively. Dashed and solid lines represent the initial and optimization relative rotation results, respectively.}
    \label{fig:gnc-ransac-test}
\end{figure}

To obtain a reasonable rotation estimation with the outlier fraction over 50\%, we proposed a fusion strategy by combining GNC and RANSAC. We set the image noise to $1$ pixel, chose $n_{s} = 30$, and conducted tests in both normal and planar scenarios for $n = 300$. The simulation results are illustrated in  \cref{fig:gnc-ransac-test}. For outlier fractions from 50\% to 80\%, our strategy consistently ensured high-accuracy rotation results. Remarkably, at a 80\% outlier fraction, it significantly outperformed other strategies in both normal and planar scenarios, with an accuracy improvement of nearly two orders of magnitude. At lower outlier fractions, our strategy maintained a considerable lead in rotation accuracy. Note that the rotation accuracy of robust-Npt is not consistent with the reported result in~\cite{zhao2020efficient} because of a simulation issue of chirality therein. Please see Appendix.B for details.

\subsection{Real Data Experiment with Strecha}


For each Strecha data, we obtained raw pair matches between two views by using SIFT~\cite{lowe2004distinctive} and the Cascade hashing~\cite{cheng2014fast} for feature extraction and matching, respectively. Then, we only processed those raw point pairs that is larger than 30 in number. Specifically, relative rotations between i-th and j-th views, namely $\hat{R}_{i,j}$, were obtained by using OpenGV's RANSAC and GNC-RANSAC schemes. The angular distance between these estimates and the true relative rotations $R_{ij}$ was computed by
\begin{equation}
\epsilon_{i,j} = \arccos \left(\frac{\operatorname{trace}\left({R}_{ {i,j}}^{\top} {\hat R_{i,j}}\right) - 1}{2}\right).
\end{equation}

\begin{figure}
    \centering
    \begin{subfigure}{0.4\linewidth}
        \centering
        \includegraphics[width=\linewidth]{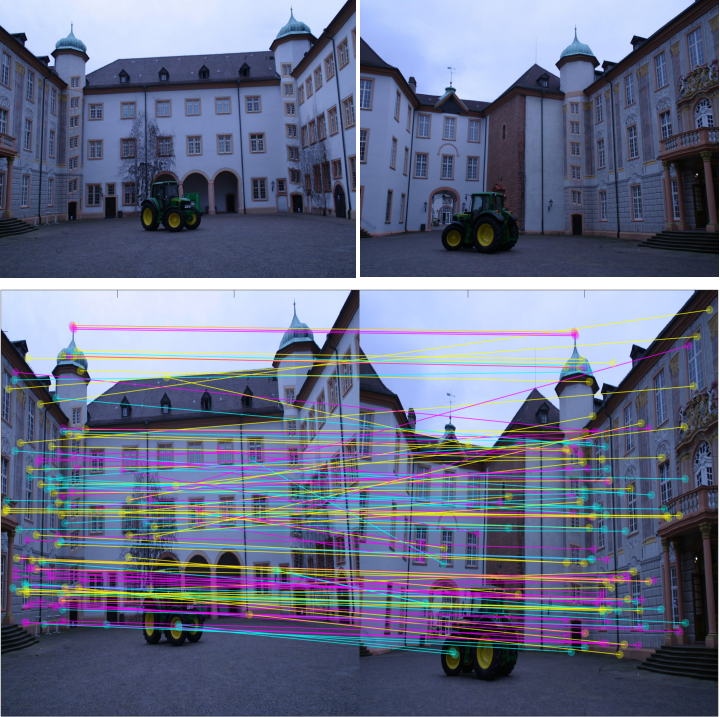}
        \captionsetup{justification=centering}
        \caption{}
        \label{fig:repeated}
    \end{subfigure}\hfill 
    \begin{subfigure}[b]{0.55\linewidth} 
        \centering
        \includegraphics[width=\linewidth]{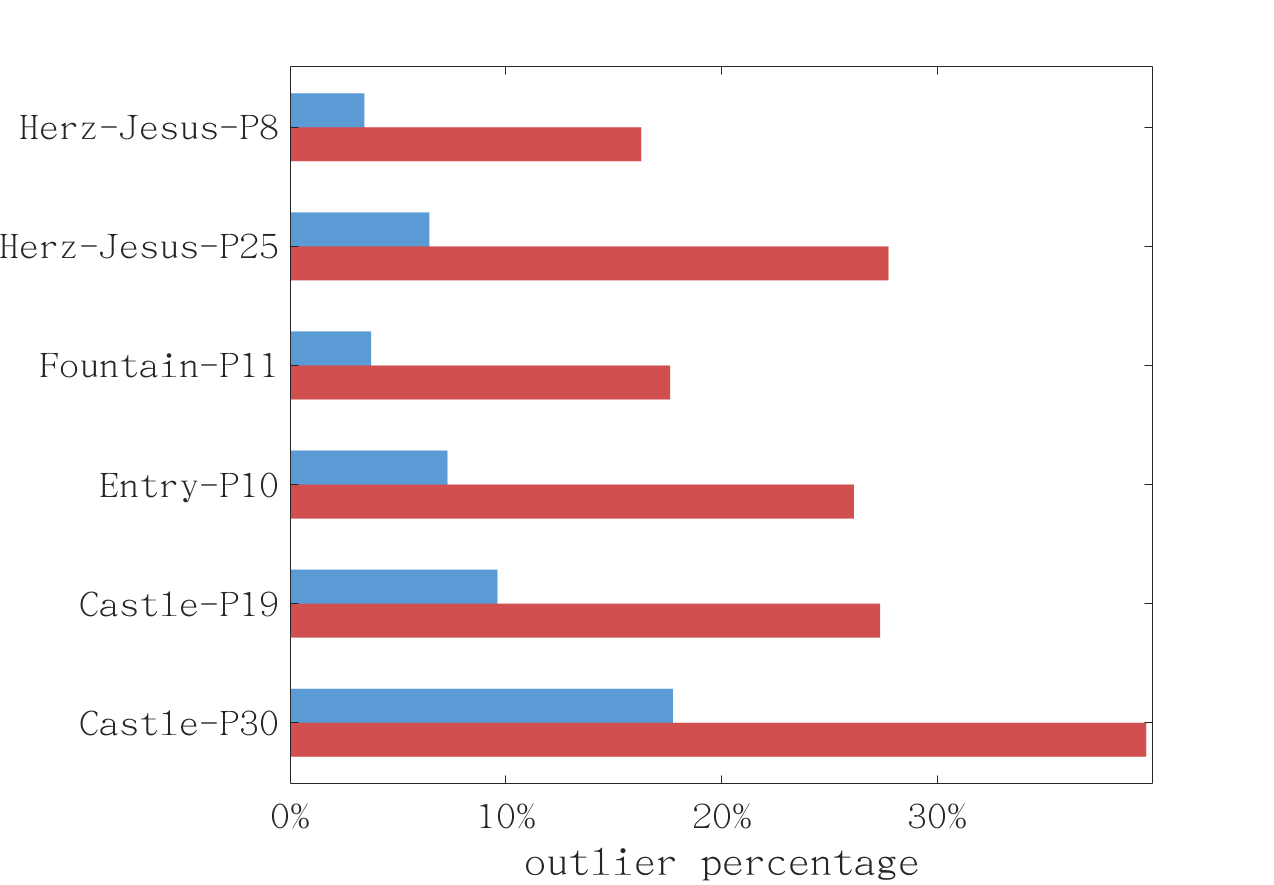}
        \captionsetup{justification=centering}
        \caption{}
        \label{fig:outlier ratios in strecha}
    \end{subfigure}
    \caption{(a) Demo of repeated texture and the (point/view) pair outliers in Castle-P30. (b) Identified outlier percentage for Strecha data (blue for OpenGV, red for GNC-RANSAC).}
    \label{fig:Strecha1}
\end{figure}

Regarding relative rotation accuracy in \cref{tab:comparison_table1}, the GNC-RANSAC method generally is superior to OpenGV by 2 $\sim$ 3 times across the Strecha dataset. Particularly, the GNC-RANSAC exhibits a median error $\epsilon_{med}$ of 0.044 for Herz-Jesus-P8, which is notably lower than OpenGV's 0.096. However, for Castle-P30, there exists a significant number of mismatches because of repeated texture, as shown in \cref{fig:repeated}. Although the proposed GNC-RANSAC method can handle a great portion of those mismatches (see \cref{fig:outlier ratios in strecha}), there are still considerable percentage wrongly taken as inliers in Castle-P30 that results in relatively larger errors than other data. \cref{fig:outlier ratios in strecha} compared the outlier percentage for all Strecha data identified by OpenGV and GNC-RANSAC schemes. The number of identified outliers by GNC-RANSAC is 2 $\sim$ 4 times of that by OpenGV, which partially explains the higher rotation accuracy of the former.

\begin{figure}
\centering
    \begin{subfigure}{0.8\linewidth}
        \includegraphics[width=\linewidth]{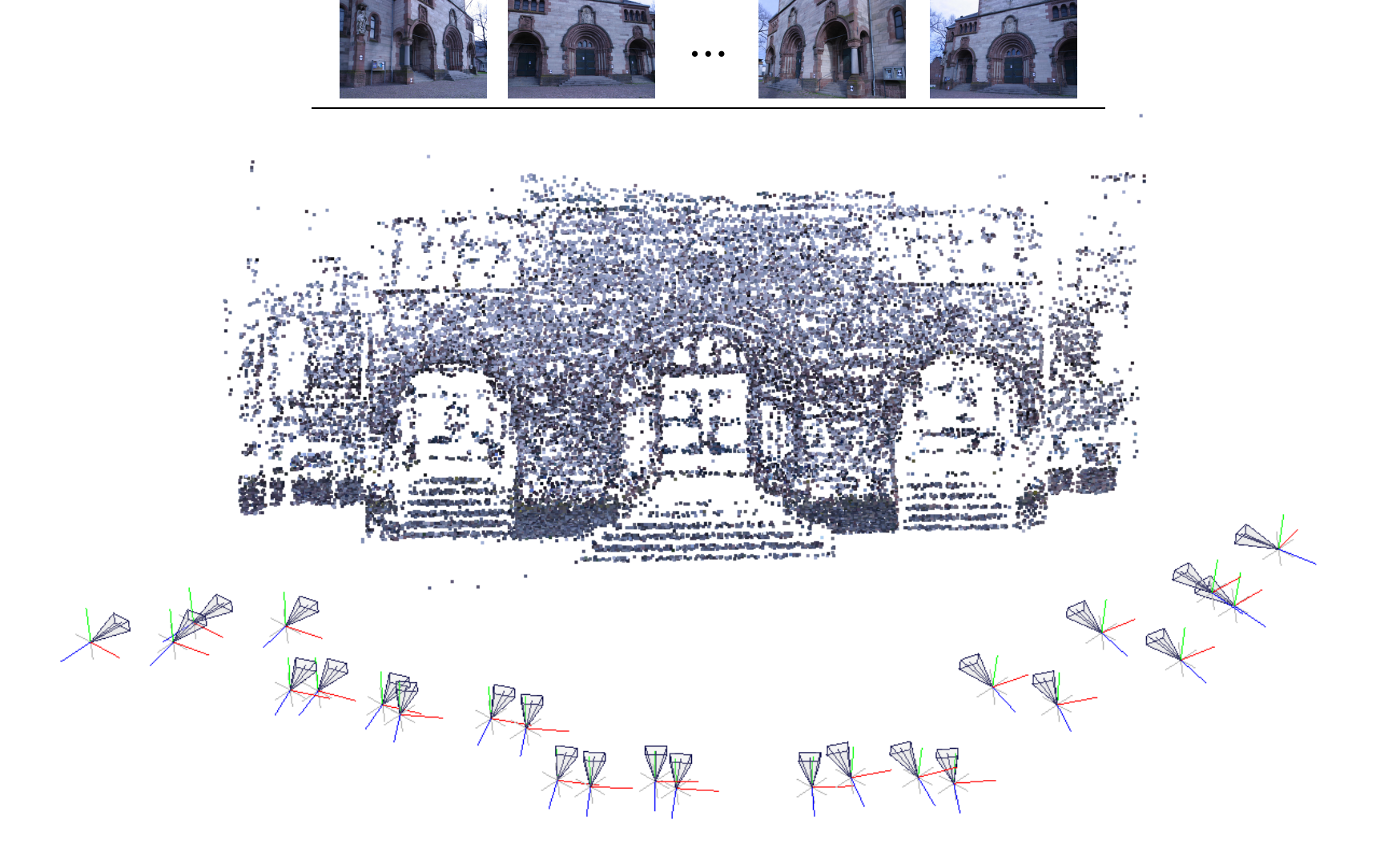}
        \captionsetup{justification=centering}
    \end{subfigure}
    \caption{Global camera poses and 3D points in Herz-Jesus-P25 obtained by the pipeline of global rotation averaging, global translation estimation, and analytical reconstruction.}
    \label{fig:poses_3DHerz}
\end{figure}

Given the relative rotations, \cref{fig:poses_3DHerz} exemplifies the global poses and 3D points of Herz-Jesus-P25 reconstructed by the pipeline of Chatterjee's global rotation averaging~\cite{chatterjee2017robust}, LiGT for global translation, and then analytical reconstruction~\cite{cai2021pose}. The apparently high quality of 3D reconstruction, at the absence of global optimization, indirectly verifies the satisfying relative rotation accuracy.

\begin{table}
    \centering
    \caption{Relative rotation accuracy (degree) comparison.}
    \label{tab:comparison_table1}
    \small 
    \setlength{\tabcolsep}{4pt} 
    \begin{tabular}{|c|c|c|c|c|}
    \hline
    Data & \multicolumn{2}{c|}{OpenGV} & \multicolumn{2}{c|}{GNC-RANSAC} \\
    \cline{2-5}
    & $\epsilon_{mean}$ & $\epsilon_{med}$ & $\epsilon_{mean}$ & $\epsilon_{med}$ \\
    \hline
    Herz-Jesus-P8 & 0.125 & 0.096 & \textbf{0.072} & \textbf {0.044} \\
    Herz-Jesus-P25 & 0.364 & 0.162 & \textbf{0.214} & \textbf{0.078} \\
    Fountain-P11 & 0.149 & 0.101 & \textbf{0.0647} & \textbf{0.0480} \\
    Entry-P10 & 0.748 & 0.180 & \textbf{0.0412} &\textbf{ 0.0367} \\
    Castle-P19 & 2.535 & 0.468 & \textbf{0.449} & \textbf{0.100} \\
    Castle-P30 & 2.385 & 0.365 & \textbf{1.949} & \textbf{0.089} \\
    \hline
    \end{tabular}
\end{table}

\section{Conclusions}
\label{sec:conclusion}
This paper introduces a linear relative pose estimation algorithm to handle planar degenerate scenes
for $n\geq6$ point pairs. The proposed algorithm utilizes the pose-only constraint to construct the residual of identifying the image matching outliers, integrated with GNC-IRLS and RANSAC for improved robustness.
Simulations and real tests of the Strecha dataset show that the proposed algorithm achieves relative rotation accuracy improvement of 2 $\sim$ 10 times in face of as large as 80\% outliers. Our future work will focus on the strategy of refining/replacing IRLS and RANSAC, for example by removing the manually-tuned parameters.


\appendix
\section{Equivalence Proof of Kneip's and Essential Equations}
\label{sec:Equivalence Proof}

Here, we provide an equivalence proof for Kneip's constraint in \cref{eq:eq6} and essential equation.

\noindent  \textbf{Proposition}: In two-view geometry, the Kneip's constraint is equivalent to the essential equation.\\
\noindent \textbf{Proof}:\\
\noindent a)	Sufficiency Proof:
Assume a matrix $B=\left(\boldsymbol{x}_1^{\prime} \times R \boldsymbol{x}_1, \ldots, \boldsymbol{x}_n^{\prime} \times R \boldsymbol{x}_n\right)^T \triangleq\left(\boldsymbol{b}_1, \ldots, \boldsymbol{b}_n\right)^T$, where
\begin{equation}
\boldsymbol{b}_i^T=\operatorname{vec}(R)^T\left(x_i\left[\boldsymbol{x}_i^{\prime}\right]_{\times} \quad y_i\left[\boldsymbol{x}_i^{\prime}\right]_{\times} \quad\left[\boldsymbol{x}_i^{\prime}\right]_{\times}\right)^T.
\label{eq:sup1}
\end{equation}
The operator $\operatorname{vec}()$ denotes the vectorization of a matrix. Let $R=\left(\begin{array}{ccc}\boldsymbol{r}_1 & \boldsymbol{r}_2 & \boldsymbol{r}_3\end{array}\right)$ and substituting it into the above equation

\begin{equation}
\begin{aligned}
\boldsymbol{b}_i^T & =-x_i \boldsymbol{r}_1^T\left[\boldsymbol{x}_i^{\prime}\right]_{\times}-y_i \boldsymbol{r}_2^T\left[\boldsymbol{x}_i^{\prime}\right]_{\times}-\boldsymbol{r}_3^T\left[\boldsymbol{x}_i^{\prime}\right]_{\times} \\
& =\left(\begin{array}{ccc}
x_i \boldsymbol{x}_i^{\prime T} & y_i \boldsymbol{x}_i^{\prime T} & \boldsymbol{x}_i^{\prime T}
\end{array}\right)
\left(\begin{array}{c}
{\left[\boldsymbol{r}_1\right]_{\times}} \\
{\left[\boldsymbol{r}_2\right]_{\times}} \\
{\left[\boldsymbol{r}_3\right]_{\times}}
\end{array}\right) \\
& \triangleq\left(\boldsymbol{x}_i^T \otimes \boldsymbol{x}_i^{\prime T}\right) G \\
& \triangleq \boldsymbol{a}_i^T G.
\label{eq:sup2}
\end{aligned}
\end{equation}
According to the Kneip's constraint: only the correct $R$ will cause the matrix $B$ to be rank deficient. That is to say, only the correct rotation matrix $R$ results in a non-zero vector $\boldsymbol{k}=\left(k_1, k_2, k_3\right)^T \in \mathbb{R}^{3 \times 1}$ satisfying
\begin{equation}
\boldsymbol{a}_i^T G \boldsymbol{k}={0}.
\label{eq:sup3}
\end{equation}
Considering the specific form of $G$ matrix, we have
\begin{equation}
-G \boldsymbol{k}=-\left(\begin{array}{c}
{\left[\boldsymbol{r}_1\right]_{\times}} \\
{\left[\boldsymbol{r}_2\right]_{\times}} \\
{\left[\boldsymbol{r}_3\right]_{\times}}
\end{array}\right) \boldsymbol{k}=\left(\begin{array}{c}
{[\boldsymbol{k}]_{\times} \boldsymbol{r}_1} \\
{[\boldsymbol{k}]_{\times} \boldsymbol{r}_2} \\
{[\boldsymbol{k}]_{\times} \boldsymbol{r}_3}
\end{array}\right) \triangleq \operatorname{vec}(Q),
\label{eq:sup4}
\end{equation}
where $Q=\left([\boldsymbol{k}]_{\times} \boldsymbol{r}_1 \quad[\boldsymbol{k}]_{\times} \boldsymbol{r}_2 \quad[\boldsymbol{k}]_{\times} \boldsymbol{r}_3\right)=[\boldsymbol{k}]_{\times} R$. Therefore,
\begin{equation}
\left(\boldsymbol{x}_i^T \otimes \boldsymbol{x}_i^{\prime T}\right) \operatorname{vec}(Q)=\boldsymbol{x}_i^{\prime T}[\boldsymbol{k}]_{\times} R \boldsymbol{x}_i={0}.
\label{eq:sup5}
\end{equation}
Substituting \cref{eq:eq2} into the above, we have
\begin{equation}
\left(R \boldsymbol{x}_i+s_i \boldsymbol{t}\right)^T[\boldsymbol{k}]_{\times} R \boldsymbol{x}_i=\boldsymbol{t}^T[\boldsymbol{k}]_{\times} R \boldsymbol{x}_i={0}.
\end{equation}
Since the above equation holds for any normalized image coordinates $\boldsymbol x_i$, it indicates $\boldsymbol k= \boldsymbol t$ (up to a scale), hence $\boldsymbol{x}_i^{\prime T} E \boldsymbol{x}_i={0}$ is valid.

\noindent b)	Necessity Proof: According to the essential equation in \cref{eq:eq4}, we have $\left(\boldsymbol{x}_i^T \otimes \boldsymbol{x}_i^{\prime T}\right) \operatorname{vec}(Q)={0}$, where $Q $ denotes the solution to the essential equation. According to~\cite{cai2019equivalent} and considering \cref{eq:eq2}, 
\begin{equation}
\left(\boldsymbol{x}_i^T \otimes\left(\begin{array}{ll}
\boldsymbol{x}_i^T \quad s_i
\end{array}\right)\right)\left(I_3 \otimes(R \quad \boldsymbol{t})^T\right) \operatorname{vec}(Q)=0
\end{equation}
For all point pairs, we have
\begin{equation}
     L \boldsymbol y = \boldsymbol 0
    \label{eq:supLy=0}
\end{equation}
where
\begin{equation}
\quad L \triangleq \left( \begin{array}{c}
\boldsymbol{x}_1^T \otimes \left( \boldsymbol{x}_1^T \quad {s}_1 \right) \\
\vdots \\
\boldsymbol{x}_n^T \otimes \left( \boldsymbol{x}_n^T \quad {s}_n \right)
\end{array} \right),
\end{equation}
and 
\begin{equation}
    \quad \boldsymbol y \triangleq \left( I_3 \otimes ({R} \quad \boldsymbol{t})^T \right) \text{vec}({Q}).
\end{equation}
Expanding the $L$ matrix, we have
\begin{equation}
\scriptsize
\setlength{\arraycolsep}{2pt}
L=\left( \begin{array}{@{}*{12}{c}@{}}
x_1^2 & x_1y_1 & x_1 & x_1s_1 & y_1x_1 & y_1^2 & y_1 & y_1s_1 & x_1 & y_1 & 1 & s_1 \\
\vdots & \vdots & \vdots & \vdots & \vdots & \vdots & \vdots & \vdots & \vdots & \vdots & \vdots & \vdots \\
x_n^2 & x_ny_n & x_n & x_ns_n & y_nx_n & y_n^2 & y_n & y_ns_n & x_n & y_n & 1 & s_n
\end{array} \right).
\end{equation}
As columns 2, 3, and 7 are respectively equal to 5, 9, and 10, we have $\text{rank}(L) \leq 9$. Let $A = L \left( I_3 \otimes (R \quad \boldsymbol t)^T \right)$, we have $\text{rank}(A) \leq 9$. Therefore, when $n \geq 9$ the homogeneous equation in \cref{eq:supLy=0} has three linearly independent special solutions
\begin{equation}
\begin{aligned}
\boldsymbol \xi_1 &= (0~1~0~0~-1~0~0~0~0~0~0~0)^T ,\\
\boldsymbol \xi_2 &= (0~0~1~0~0~0~0~0~-1~0~0~0)^T, \\
\boldsymbol \xi_3 &= (0~0~0~0~0~0~1~0~0~-1~0~0)^T.
\end{aligned}
\end{equation}
The solution space of $\boldsymbol y $ is given by
\begin{equation}
\boldsymbol y = (I_3 \otimes (R \quad \boldsymbol t))^T \text{vec}(Q) = \begin{pmatrix} \boldsymbol \xi_1 & \boldsymbol \xi_2 & \boldsymbol \xi_3 \end{pmatrix} \boldsymbol c,
\end{equation}
where $\boldsymbol c = (c_1,c_2,c_3)^T$ is a non-zero real vector. By using the Kronecker
product equality
\begin{equation}
    (I_3 \otimes (R \quad \boldsymbol t))^T \text{vec}(Q) = \text{vec}((R \quad \boldsymbol t)^T Q),
\end{equation}
we can obtain
\begin{equation}
    Q = R[\boldsymbol c]_{\times} = [ \boldsymbol k]_{\times} R,
\end{equation}
where $\boldsymbol k = R \boldsymbol c$. It  means that there exists at least one non-zero vector $\boldsymbol{k}=\left(k_1, k_2, k_3\right)^T \in \mathbb{R}^{3 \times 1}$ satisfying $Q=[\boldsymbol{k}]_{\times} R$, which is $\operatorname{vec}(Q)=G \boldsymbol{k}$. Substituting into the essential equation in \cref{eq:eq4}, we have $\boldsymbol{a}_i^T G \boldsymbol{k}=\boldsymbol{b}_i^T \boldsymbol{k}=0$, i.e., matrix $B$ must be rank deficient, and hence the essential equation implies the Kneip constraint.

In summary, the Kneip's constraint and the essential equation are equivalent. Q.E.D.

\section{Performance of Robust N-point Method}
\label{sec:performance_robust_npt}
The rotational accuracy of the robust N-point method, as presented in \cref{sec:Experiments}, does not align with the results reported by Zhao~\cite{zhao2020efficient}. In our robustness experiments, we employed 300 point pairs satisfying the chirality constraint, with each pair generated at a noise level of 1 pixel, subject to Gaussian distribution. We fixed the maximum translation magnitude at 2 units and had the depth of 3D points ranged from 4 to 18. An attitude perturbation of 0.5 degrees was applied. To ensure reliability, we repeated the experiment 500 times. For the RANSAC algorithm, we used the default OpenGV settings.

\begin{figure}
    \centering
    \begin{subfigure}{0.5\linewidth}
        \includegraphics[width=\linewidth]{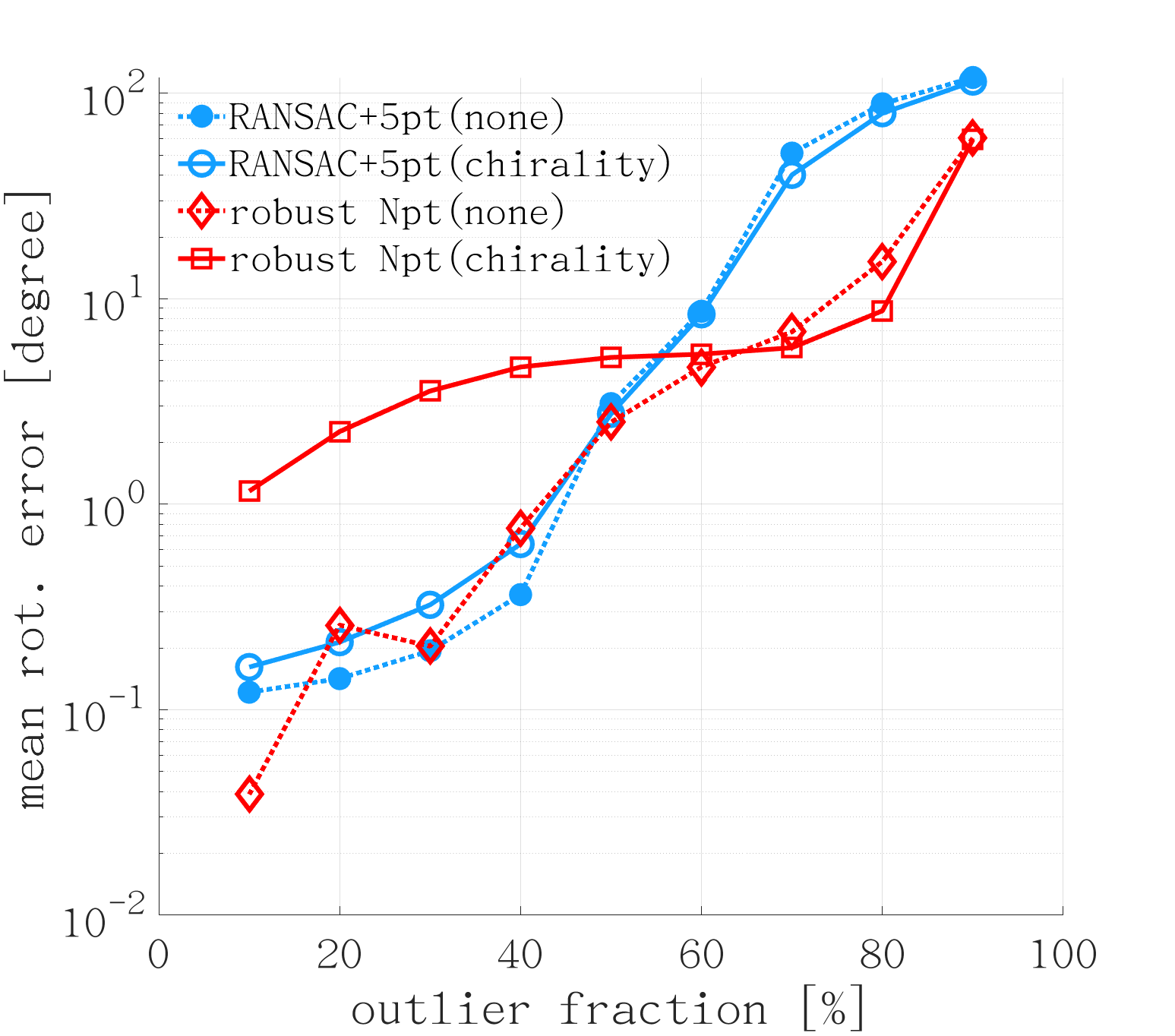}
        \caption{}
        \label{fig:removed_chirality_rnpt_mean}
    \end{subfigure}%
    \begin{subfigure}{0.5\linewidth}
        \includegraphics[width=\linewidth]{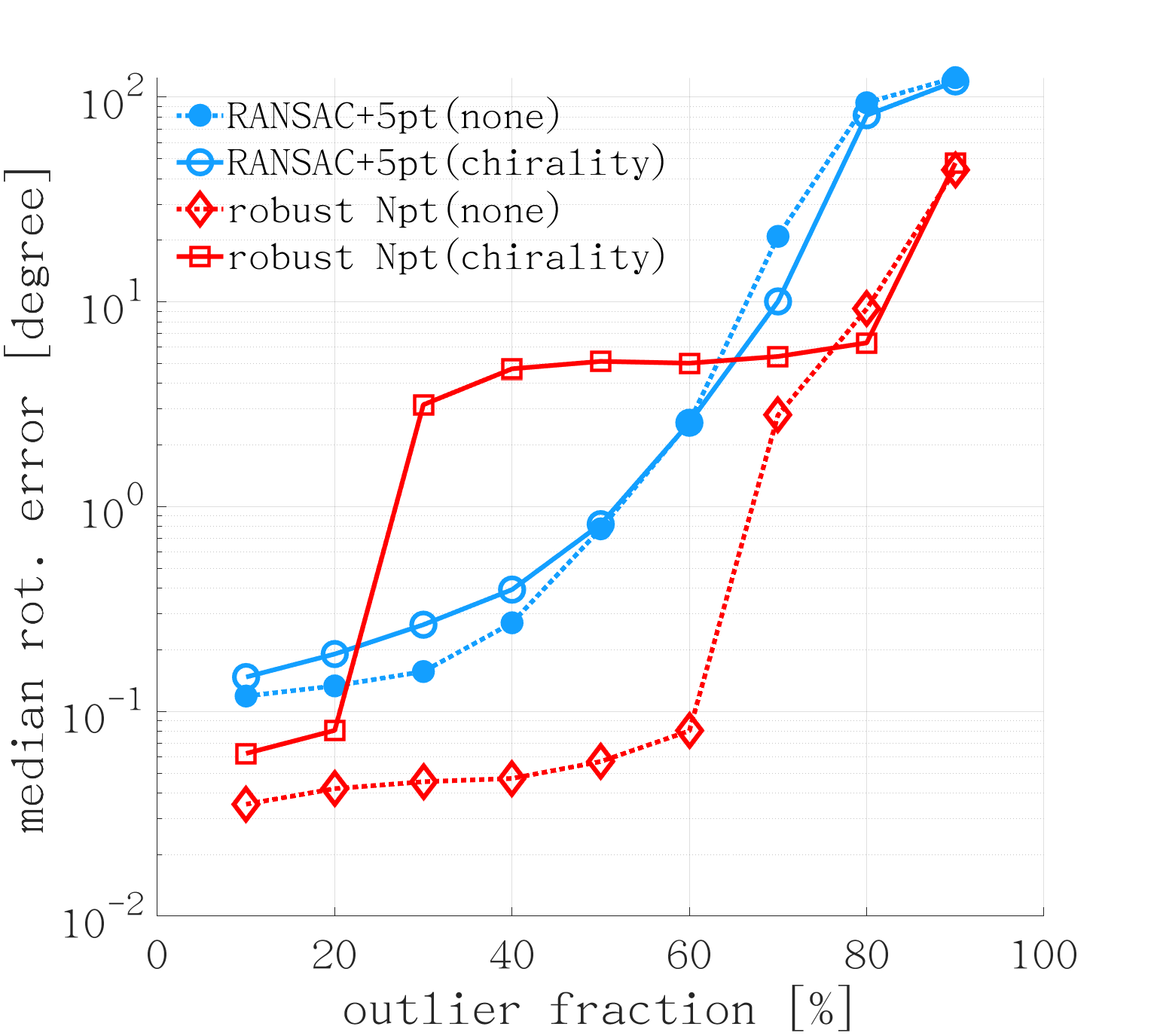}
        \caption{}
        \label{fig:removed_chirality_rnpt_median}
    \end{subfigure}
    \caption{Performance of robust Npt and RANSAC+5pt in cases of disregarding or considering the chirality constraint. We use~\textquotesingle none\textquotesingle~to denote the point pair generation disregarding the chirality constraint, and~\textquotesingle chirality\textquotesingle~to represent the generation considering the chirality constraint.}
    \label{fig:removed_chirality_rnpt}
\end{figure}

In contrast, the point pairs used in the robust N-point method~\cite{zhao2020efficient} were generated from OpenGV's simulations that did not consider the chirality constraint. Thus, we compared the performance of robust N-point method in cases of disregarding or considering the chirality constraint in \cref{fig:removed_chirality_rnpt}. In the case of disregarding the chirality constraint, the mean and median errors of the robust N-point method significantly reduced. In other words, the results reported in~\cite{zhao2020efficient} are not objective in that those simulated point pairs disregarding the chirality constraint should not have been taken into account. Finally, in cases with 90\% outlier fraction, as depicted in \cref{fig:removed_chirality_rnpt}, the rotation errors of OpenGV's RANSAC+5pt method are around $10^2$ degrees, which also contrasts with the result of 10 degrees in~\cite{zhao2020efficient}.

{
    \small
    \bibliographystyle{ieeenat_fullname}
    \bibliography{main}
}


\end{document}